\documentclass[11pt,a4paper]{article}
\usepackage[]{coling}
\usepackage[colorlinks=true,linkcolor=black,anchorcolor=gray,urlcolor=black,citecolor=gray]{hyperref}
\usepackage{times}
\usepackage{graphicx}
\usepackage{latexsym}
\usepackage{tabulary,booktabs,tabularx}
\usepackage{url}
\usepackage{listings}
\usepackage{comment,wrapfig,adjustbox}
\usepackage[table,xcdraw]{xcolor}
\usepackage[flushleft]{threeparttable}
\usepackage{cleveref}
\usepackage{adjustbox}
\usepackage[normalem]{ulem}
\usepackage{ragged2e}
\usepackage{hyperref}
\usepackage{textcomp}
\usepackage{amsfonts}  %
\usepackage{placeins}  %
\usepackage{subcaption}
\usepackage{xcolor}

\usepackage{tablefootnote}

\definecolor{codegreen}{rgb}{0,0.6,0}
\definecolor{codegray}{rgb}{0.5,0.5,0.5}
\definecolor{codepurple}{rgb}{0.58,0,0.82}
\definecolor{backcolour}{rgb}{0.95,0.95,0.92}

\usepackage{float}
\floatstyle{plain} %
\newfloat{Code}{!t}{myc}
\lstdefinestyle{mystyle}{
    backgroundcolor=\color{backcolour},   
    commentstyle=\color{codegreen},
    keywordstyle=\color{magenta},
    numberstyle=\tiny\color{codegray},
    stringstyle=\color{codepurple},
    basicstyle=\ttfamily\scriptsize,
    breakatwhitespace=false,         
    breaklines=true,                 
    captionpos=b,                    
    keepspaces=true,                 
    numbers=none,                    
    numbersep=5pt,                  
    showspaces=false,                
    showstringspaces=false,
    showtabs=false,                  
    tabsize=2
}

\newenvironment{itemize*}%
  {\begin{itemize}%
    \setlength{\itemsep}{.2pt}%
    \setlength{\parskip}{.2pt}%
    \setlength{\topsep}{.5pt}}%
  {\end{itemize}}

\colingfinalcopy

\title{\name: Dense Temporal Annotation on a Timeline}

\author{
    Anna Rogers \\ Dept. of Social Data Science \\ University of Copenhagen \\ \texttt{\small arogers@sodas.ku.dk} \\ 
    \And Marzena Karpinska \\ Dept. of Computer Science \\ UMass Amherst \\ \texttt{\small mkarpinska@cs.umass.edu} \\ 
    \And Ankita Gupta\\ Dept. of Computer Science \\ UMass Amherst \\ \texttt{\small ankitagupta@cs.umass.edu} \\ 
    \AND Vladislav Lialin\\ Dept. of Computer Science \\ UMass Lowell \\ \texttt{\small vlialin@cs.uml.edu}\\ 
    \And Gregory Smelkov\\ Dept. of Computer Science \\ UMass Lowell \\ \texttt{\small gsmelkov@cs.uml.edu} \\ 
    \And Anna Rumshisky\\  Dept. of Computer Science \\ UMass Lowell \\ \texttt{\small arum@cs.uml.edu}\\
}
\newif\ifcomment
\commenttrue %
\ifcomment

\newcommand{\arucomment}[1]{\textcolor{red}{\bf \small [ #1 --ARU]}}
\definecolor{atomictangerine}{rgb}{1.0, 0.3, 0.8}
\newcommand{\mkcomment}[1]{\textcolor{atomictangerine}{\bf \small [ #1 --MK]}}

\else
\newcommand{\arocomment}[1]{}
\newcommand{\arucomment}[1]{}
\newcommand{\mkcomment}[1]{}
\fi

\newcommand{\name}[0]{\textsc{NarrativeTime}}
\newcommand{\tlinks}[0]{\textsc{tlinks}}
\newcommand{\tlink}[0]{\textsc{tlink}}

\date{}

\begin{document}
\maketitle

\begin{abstract}

For the past decade, temporal annotation has been sparse: only a small portion of event pairs in a text was annotated. We present \name, the first timeline-based annotation framework that achieves full coverage of all possible \tlinks. To compare with the previous SOTA in dense temporal annotation, we perform full re-annotation of TimeBankDense corpus, which shows comparable agreement with a signigicant increase in density. We contribute TimeBankNT corpus (with each text fully annotated by two expert annotators), extensive annotation guidelines, open-source tools for annotation and conversion to TimeML format, baseline results, as well as quantitative and qualitative analysis of inter-annotator agreement.

\end{abstract}

\section{Introduction}

\begin{figure}[b]
    \centering
    \includegraphics[width=.9\textwidth]{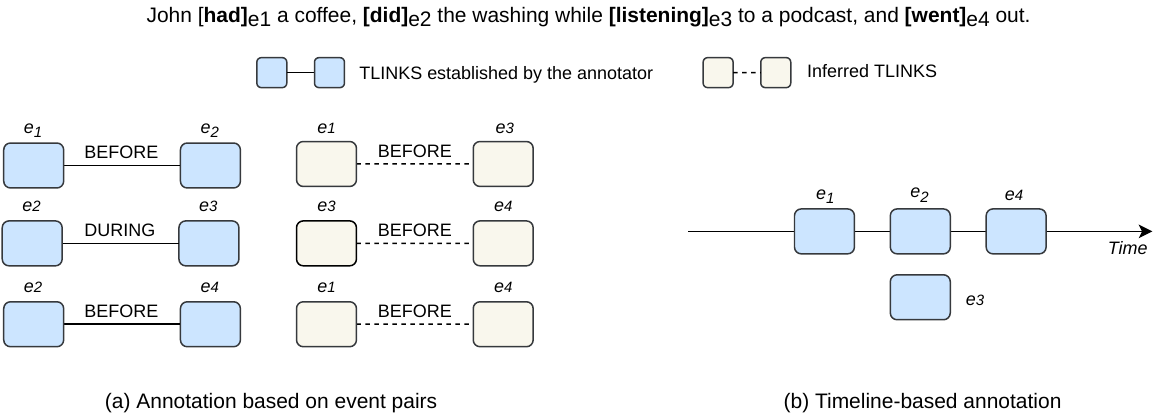}
    \caption{Timeline-based annotation vs annotation based on event pairs.}
    \label{fig:comparison}
\end{figure}

Event order information is usually represented by  temporal links (\tlinks) between events pairs: does $event_1$ happen \textsc{Before}/\textsc{During}/\textsc{After} $event_2$? Ideally, temporal annotation would establish \textit{all} \tlinks\ in the text. However, their number is quadratic to the number of events in the text, so temporal annotation is usually \textit{sparse}: e.g. TimeBank only contains 1-5\% of all possible \tlinks\ \cite{Verhagen_2005_Temporal_closure_in_annotation_environment}. %
Furthermore, much of this information is underspecified in the text, and is not normally inferred by human readers (nor do they make the same inferences if pressed to do so). Several solutions have been proposed for the density problem \cite{Verhagen_2005_Temporal_closure_in_annotation_environment,CassidyMcDowellEtAl_2014_An_Annotation_Framework_for_Dense_Event_Ordering} and for the underspecification problem \cite{BethardKolomiyetsEtAl_2012_Annotating_Story_Timelines_as_Temporal_Dependency_Structures,NingWuEtAl_2018_A_Multi-Axis_Annotation_Scheme_for_Event_Temporal_Relations}, but they remain a challenge.

We address both of these problems in \name, the first \textit{timeline-based framework for full temporal annotation}. While the traditional TimeBank-style annotation focuses on relations in individual event pairs, partly annotated and partly inferred (\autoref{fig:comparison}a), \name\ builds a dynamic  timeline (\autoref{fig:comparison}b). That representation is equivalent to the full set of all possible \tlinks\ in the text, and they are guaranteed to be backed by manual annotation (which may not be the case for the pairwise approach). %
Its solutions to the underspecification problem is based on three mechanisms: event types, timeline branches and factuality.

We implement \name\ framework in detailed annotation guidelines and open-source tools\footnote{\url{https://github.com/text-machine-lab/nt}} for annotation and conversion to the standard TimeML format. To enable direct comparison between  our approach and prior work, we re-annotate the  TimeBank-Dense \cite{CassidyMcDowellEtAl_2014_An_Annotation_Framework_for_Dense_Event_Ordering} corpus. We achieve inter-annotator agreement (IAA) of Krippendorff's $\alpha$ 0.68 \cite{krippendorff2004content}, which can be interpreted as a substantial agreement \cite{Landis1977,artstein-poesio-iaa-2008}. %
This is comparable or superior to what is reported in most sparse annotation projects, but \name\ annotation is dense: it yields 102,313 \tlinks\footnote{TimeBankNT contains 102,313 \tlinks\ excluding inverses (symmetrical \tlinks\ that can be auto-inferred, such as X \textsc{Before} Y $\rightarrow$ Y \textsc{After} X), and 204,626 \tlinks\ including inverses.} vs 12,715 \tlinks\ in the original TimeBankDense \cite{CassidyMcDowellEtAl_2014_An_Annotation_Framework_for_Dense_Event_Ordering} and 1,341 \tlinks\ in the same files in the original TimeBank \cite{PustejovskyHanksEtAl_2003_The_TIMEBANK_Corpus}. 

In our study, each TimeBank-Dense text was independently and fully annotated by two expert annotators. Since temporal annotation has many cases of genuine variation in human interpretation, where a single ``ground truth'' is unrealistic, such disaggregated data is needed for ``learning from disagreements'' \cite{UmaFornaciariEtAl_2021_Learning_from_Disagreement_Survey,Plank_2022_Problem_of_Human_Label_Variation_On_Ground_Truth_in_Data_Modeling_and_Evaluation}. We perform qualitative and quantitative analysis of inter-annotator agreement and label variation, %
and we release the  annotated data. We also contribute initial benchmark results, based on LongT5 \cite{Guo2021LongT5ET} Transformer encoder.

To clarify the terminology: we use the term \textit{framework} to differentiate between annotation workflows that are based on relations between individual event pairs, and timeline-based annotation. \textit{Annotation scheme} refers to the specific set of policies about what to annotate and how, which is implemented in \textit{annotation guidelines}. Both timeline- and event-pair-based frameworks can support different annotation schemes. The results of annotation in either framework can be represented in ISO-TimeML \textit{format} 
\cite{PustejovskyLeeEtAl_2010_ISO-TimeML_An_International_Standard_for_Semantic_Annotation}
encoded as as a collection of \tlinks\ between event pairs.

\begin{table}[!t]
\footnotesize
\begin{threeparttable}
\begin{adjustbox}{max width=\textwidth}
\begin{tabular}{p{4.6cm}p{.6cm}p{.6cm}p{.7cm}p{1.25cm}p{1.2cm}p{.8cm}p{.8cm}p{1cm}%
}
\toprule
Annotation scheme & TLink types & Events IAA & TLinks IAA & TLink type IAA & IAA Metric & Corpus genre & Num. events & Num. TLinks \\ %
\midrule
TimeML %
\cite{PustejovskyIngriaEtAl_2005_specification_language_TimeML,PustejovskyLeeEtAl_2010_ISO-TimeML_An_International_Standard_for_Semantic_Annotation} & 13 & 0.78 & n/a & 0.55 & AvgPnR & news & 7,935 & 3,481 \\
\addlinespace[0.5em]
TempEval-1 \cite{VerhagenGaizauskasEtAl_2007_SemEval-2007_Task_15_TempEval_Temporal_Relation_Identification,VerhagenGaizauskasEtAl_2009_The_TempEval_challenge_identifying_temporal_relations_in_text} & 6 & n/a & n/a & 0.47  & Cohen $\kappa$ & news & 7,935 & 2,002 \\
\addlinespace[0.5em]
TempEval-3 \cite{UzZamanLlorensEtAl_2012_Tempeval3_Evaluating_events_time_expressions_and_temporal_relations} & 13 & 0.87 & n/a & n/a  & F1 & web & 11,145 &  11,098 \\
\addlinespace[0.5em]
THYME-TimeML  \cite{StylerBethardEtAl_2014_Temporal_Annotation_in_the_Clinical_Domain} & 5 & 0.79 & 0.50 & 0.50 &  Krippend-orff $\alpha$ & clinical & 15,769 & 7,935 \\
\addlinespace[0.5em]
Temporal Dependency Structure  \cite{KolomiyetsBethardEtAl_2012_Extracting_Narrative_Timelines_as_Temporal_Dependency_Structures,BethardKolomiyetsEtAl_2012_Annotating_Story_Timelines_as_Temporal_Dependency_Structures} & 6 & 0.86 & 0.82 & 0.7  & Krippend-orff $\alpha$ & fables &  1,233 & 1,139 \\
\addlinespace[0.5em]
MATRES  \cite{NingWuEtAl_2018_A_Multi-Axis_Annotation_Scheme_for_Event_Temporal_Relations} & 4 & 0.85 & n/a & 0.84\tnote{1} %
& Cohen $\kappa$ & news & 6,099%
&  13,577\tnote{2} %
\\
\addlinespace[0.5em]
RED  \cite{OGormanWright-BettnerEtAl_2016_Richer_Event_Description_Integrating_event_coreference_with_temporal_causal_and_bridging_annotation,IkutaStylerEtAl_2014_Challenges_of_adding_causation_to_richer_event_descriptions} & 4 & 0.86 & 0.73 & 0.18-0.54 & F1 & news & 8,731 & 4,969 \\
\addlinespace[0.5em]
TimeBank-Dense \cite{CassidyMcDowellEtAl_2014_An_Annotation_Framework_for_Dense_Event_Ordering} & 6 & n/a & n/a & 0.56-0.64 & Cohen $\kappa$ & news & 1,729 & 12,715 \\
\addlinespace[0.5em]
NewsReader \cite{MinardSperanzaEtAl_2016_MEANTIME_the_NewsReader_Multilingual_Event_and_Time_Corpus,vanErpVossenEtAl_2015_Annotated_Data_version_2} & 13 & 0.68 & n/a & n/a & Dice's coef. & news & 2,096 & 1,789 \\
\addlinespace[0.5em]
Araki et al. \cite{ArakiMulafferEtAl_2018_Interoperable_Annotation_of_Events_and_Event_Relations_across_Domains} & 2 & 0.80 (F1) & n/a & 0.11-0.14 & Fleiss $\kappa$ & simple wiki &  5,397 & 2,833 \\
\addlinespace[0.5em]
CaTeRS \cite{MostafazadehGrealishEtAl_2016_CaTeRS_Causal_and_Temporal_Relation_Scheme_for_Semantic_Annotation_of_Event_Structures} & 4 & 0.91 & n/a & 0.51 & Fleiss $\kappa$ & stories & 2,708 & 2,715 \\
\addlinespace[0.5em]
\addlinespace[0.5em]
UDS-T \cite{VashishthaVanDurmeEtAl_2019_Fine-Grained_Temporal_Relation_Extraction} & 2 & n/a & 0.67 & n/a & Spearman & web%
& 32,302 & 70,368 \\
\addlinespace[0.5em]
TDG \cite{YaoQiuEtAl_2020_Annotating_Temporal_Dependency_Graphs_via_Crowdsourcing} & 4 & 0.79 & 0.52-0.85 & 0.85-0.91 & F1 & wiki%
& 14,974 & 28,350 \\
\addlinespace[0.5em]
MAVEN-ERE \cite{maven-ere} & 6 & n/a & 0.678 & n/a & Cohen $\kappa$  & wiki%
& 103,193 & 1,216,217 \\
\addlinespace[0.5em]
\bottomrule
\end{tabular}
\end{adjustbox}
  \begin{tablenotes}
    \item[1] Both coefficients of agreement are reported for two expert annotators who annotated a small portion of data (about 100 events and 400 relations).
    \item[2] Since the initial release MATRES was extended to include the entire TempEval3 dataset (only verbal events). We cite the numbers for the newer, extended version available at \url{https://github.com/qiangning/MATRES}.
  \end{tablenotes}

\end{threeparttable}
\caption{Statistics reported in the current temporal annotation projects for English. %
}
\label{tab:related_work}
\end{table}

\section{Related work}
\label{sec:review}

To the best of our knowledge, all current proposals for temporal annotation are based on the event-pair-based framework. Within that framework, there are different annotation schemes that have been applied to different text corpora. 

A summary of major available resources is presented in \autoref{tab:related_work}, which  %
shows that the task of annotating event order is not characterized by high agreement, and there is no real consensus even on what agreement metric to use. The reported IAA for identifying events tends to be considerably higher than IAA for either establishing \tlinks, or for their type. %

One of the fundamental problems for temporal annotation is that a complete set of temporal relations in a text would be quadratic on the number of events in that text, and estblishing them all would be prohibitively labor-intensive. Therefore most of existing work limit the scope of the task: only annotating \tlinks\ in the same or adjacent sentences \cite{VerhagenGaizauskasEtAl_2007_SemEval-2007_Task_15_TempEval_Temporal_Relation_Identification,VerhagenSauriEtAl_2010_SemEval-2010_Task_13_TempEval-2,UzZamanLlorensEtAl_2012_Tempeval3_Evaluating_events_time_expressions_and_temporal_relations,MinardSperanzaEtAl_2016_MEANTIME_the_NewsReader_Multilingual_Event_and_Time_Corpus}, limiting the scope to a specific construction \cite{BethardMartinEtAl_2007_Timelines_from_Text_Identification_of_Syntactic_Temporal_Relations}. Another line of work focuses on trying to infer the missing \tlinks\ via transitive closure \cite{SetzerGaizauskas_2001_Pilot_Study_On_Annotating_Temporal_Relations_In_Text,Verhagen_2005_Temporal_closure_in_annotation_environment,ManiVerhagenEtAl_2006_Machine_learning_of_temporal_relations}. However, this process is not conflict-free \cite{Verhagen_2005_Temporal_closure_in_annotation_environment}, and the current methods to produce full temporal graphs from sparse annotations are not very successful \cite{OcalPerezEtAl_2022_Holistic_Evaluation_of_Automatic_TimeML_Annotators}. A key problem is that the existing annotations often suffice only to construct local event chains, but there is not enough information to connect them \cite{ChambersJurafsky_2008_Jointly_Combining_Implicit_Constraints_Improves_Temporal_Ordering}.

In addition to laboriousness, establishing the full set of all possible \tlinks\ is difficult because human readers %
do not seem to even infer all of these relations for every text they read. Much of this information is naturally left underspecified, and if the annotators are forced to infer such relations, their agreement would not be high. The chief solution for the underspecification problem has been to either allow sparse annotation, to introduce additional restrictions to avoid annotating non-actual events \cite{BethardKolomiyetsEtAl_2012_Annotating_Story_Timelines_as_Temporal_Dependency_Structures} or, more recently, place them on separate axes \cite{NingWuEtAl_2018_A_Multi-Axis_Annotation_Scheme_for_Event_Temporal_Relations}. 

We address both the problems of density and underspecification by developing a new annotation framework, which offers a holistic view of the narrative represented as a timeline, rather than individual event pairs. This solves the density problem: as shown in \autoref{fig:comparison}, a timeline contains all the information needed for ordering \textit{all} event pairs. It also enables a novel solution to the underspecification problem: we incorporate vagueness in the event type definitions that have different timeline visualisations, as will be discussed in \autoref{sec:etypes}. 

Since we do not directly annotate \tlinks, but a structure from which they can be unambiguously inferred, our approach resembles the annotation of temporal dependency graphs and trees \cite{KolomiyetsBethardEtAl_2012_Extracting_Narrative_Timelines_as_Temporal_Dependency_Structures,ZhangXue_2018_Structured_Interpretation_of_Temporal_Relations,ZhangXue_2019_Acquiring_Structured_Temporal_Representation_via_Crowdsourcing_Feasibility_Study,YaoQiuEtAl_2020_Annotating_Temporal_Dependency_Graphs_via_Crowdsourcing}, where the annotators establish temporal relations as child-parent relationship in a dependency tree. However, that approach has to assume a single parent-child relation, and the annotation process still requires considering individual pairs of events or events with temporal expression, while we allow for event clusters (\autoref{sec:clusters}). The dependency structure is also less amenable to express vagueness and underspecification than our timeline-based proposal. Furthermore, temporal dependency trees may be more temporally indeterminate than the TimeML annotations \cite{OcalFinlayson_2020_Evaluating_Information_Loss_in_Temporal_Dependency_Trees}.

A number of previous projects used timeline-like representations  %
\cite{VerhagenKnippenEtAl_2006_Annotation_of_Temporal_Relations_with_Tango,KolomiyetsBethardEtAl_2012_Extracting_Narrative_Timelines_as_Temporal_Dependency_Structures,DoLuEtAl_2012_Joint_Inference_for_Event_Timeline_Construction,CaselliVossen_2016_The_Storyline_Annotation_and_Representation_Scheme_StaR_A_Proposal,CaselliVossen_2017_The_Event_StoryLine_Corpus_A_New_Benchmark_for_Causal_and_Temporal_Relation_Extraction}, 
but only as a representation of the final result: the annotation itself was still based on event pairs. %
\newcite{VashishthaVanDurmeEtAl_2019_Fine-Grained_Temporal_Relation_Extraction} proposed a framework where the annotators work with only two adjacent sentences to create a mini-timeline of the events in those two sentences. This %
enables crowdsourcing, but necessarily limits the annotation to adjacent sentences (and only a subset of those, in practice). Most recently, \newcite{maven-ere} stated that they developed and used a timeline-based annotation scheme to improve annotation density, but provided no further details, tools or the guidelines with which this was achieved.  %

\section{Why event pairs are problematic: motivation in psychology}
\label{sec:cog}

The exact mechanisms of reading comprehension are still debated \cite{RaynerReichle_2010_Models_of_the_Reading_Process}, but there are good reasons to believe that we gradually build a mental model of the whole narrative \cite{vanderMeerBeyerEtAl_2002_Temporal_order_relations_in_language_comprehension,Zwaan_2016_Situation_models_mental_simulations_and_abstract_concepts_in_discourse_comprehension}. This model has a directional representation of time and temporal distance between events, and is built correctly even if the text is not organized chronologically, e.g. if there are flashbacks \cite{Claus_2012_Processing_Narrative_Texts_Melting_Frozen_Time}.

We also know that texts pre-chunked in semantically coherent segments are easier to process \cite{FraseSchwartz_1979_Typographical_cues_that_facilitate_comprehension,OSheaSindelar_1983_The_Effects_of_Segmenting_Written_Discourse_on_the_Reading_Comprehension_of_Low-_and_High-Performance_Readers,RajendranDuchowskiEtAl_2013_Effects_of_text_chunking_on_subtitling_A_quantitative_and_qualitative_examinationa}. %
For dynamic situations, ``semantic coherence'' is best explained in terms of scripts/frames, mental representations of stereotypical complex activities. They have internal organization, with possibly complex sub-elements that can be managed without losing track of the overall goal of the script %
\cite{FaragTroianiEtAl_2010_Hierarchical_Organization_of_Scripts_Converging_Evidence_from_fMRI_and_Frontotemporal_Degeneration}. %

The process of constructing a mental model of a narrative is likely to be subject to the same on-line constraints\footnote{Reading comprehension in particular is influenced by the working memory capacity \cite{SeigneuricEhrlichEtAl_2000_Working_memory_resources_and_childrens_reading_comprehension}, vocabulary proficiency \cite{QuinnWagnerEtAl_2015_Developmental_Relations_Between_Vocabulary_Knowledge_and_Reading_Comprehension_A_Latent_Change_Score_Modeling_Study}, and even individual differences in statistical learning \cite{MisyakChristiansenEtAl_2010_On-Line_Individual_Differences_in_Statistical_Learning_Predict_Language_Processing}.} as the rest of language processing. %
This brings into play the ``good-enough processing" \cite{Christianson_2016_When_language_comprehension_goes_wrong_for_the_right_reasons_Good-enough_underspecified_or_shallow_language_processing,FerreiraEngelhardtEtAl_2009_Good_Enough_Language_Processing_A_Satisficing_Approach}. Not all temporal relations \textit{can} be inferred, since the writers focus on advancing their story in an engaging way rather than spelling out all the details. The readers also have limited time and attention, and focus on salient developments with the characters, often ignoring the details. This is the fundamental reason for the underspecification problem in temporal annotation. %

Counter-intuitively, readers do \textit{not} save effort by looking at each segment only once: we regress as needed \cite{SchotterTranEtAl_2014_Dont_Believe_What_You_Read_Only_Once_Comprehension_Is_Supported_by_Regressions_During_Reading}, even across sentence boundaries  \cite{ShebilskeReid_1979_Reading_Eye_Movements_Macro-structure_and_Comprehension_Processes}. %
This suggests that during reading a good-enough representation of the narrative is constructed, with the readers anticipating the developments \cite{Coll-FloritGennari_2011_Time_in_language_event_duration_in_language_comprehension} and filling the most glaring gaps with their world knowledge. The variation is particularly notable with regards to the length of durative events \cite{Coll-FloritGennari_2011_Time_in_language_event_duration_in_language_comprehension}. This would explain the relatively low inter-annotator agreement observed in previous temporal annotation projects.

If the above view of reading comprehension is correct, it is the opposite of the process required from annotators in a schema based on event pairs. The annotators are explicitly asked about the temporal order of two events, which may or may not be in the category of events that were salient enough in the discourse to be easily order-able. Furthermore, there is no allowance for the fact that underspecified relations are not just ``vague": if they are salient enough, their order \textit{will} be inferred, but that interpretation may well be different for different annotators, since they draw on their own world knowledge (see \autoref{sec:qualitative} for examples of such cases). %

\section{\name\  framework}
\label{sec:nt}

\subsection{What counts as event}
\label{sec:event_def}

\name\  annotation is performed in two stages: (1) identification of events and their coreference, and (2) their temporal ordering. This paper focuses on our new temporal ordering strategies: as shown in \autoref{tab:related_work}, detection of events is an easier task with a relatively high IAA, and we do not introduce anything new here. Events are ``anything that happens or occurs'' %
\cite{PustejovskyCastanoEtAl_2003_TimeML_Robust_Specification_of_Event_and_Temporal_Expressions_in_Text}, expressed as verbs, nominals, adjectives/participles, or phrases. As in TimeML, states do also count as events. Since in this case study we re-annotate existing TimeBank data, we reuse the original event annotations.

\subsection{Event types}
\label{sec:etypes}

Most current temporal annotation schemes adopt a model of temporal relations based on interval algebra \cite{allen1984towards}. Start and endpoints of 2 events form %
13 possible relations: \textsc{Before/After}, \textsc{Immediately Before/After}, \textsc{Overlap/Is overlapped}, \textsc{ends/is ended on}, \textsc{starts/is started on}, \textsc{during}, and \textsc{identity}.

However, mental tracking of all the start/endpoints is psychologically unrealistic. \newcite{NingWuEtAl_2018_A_Multi-Axis_Annotation_Scheme_for_Event_Temporal_Relations} suggest focusing on start points due to variation in perceived event durations \cite{Coll-FloritGennari_2011_Time_in_language_event_duration_in_language_comprehension}, but this assumes that the start of events is always more salient than other phases. That can hardly be the case, since focus depends on contextual saliency: for example, we would be more concerned with the end of a resuscitation activity than its beginning.

We propose integrating some temporal order information in event definitions rather than leaving it all to \tlinks. The annotators need to be able to focus on the start, end, or the ongoing phase of an event, or any combination thereof that is salient in the context, and leave out the underspecified parts. This idea owes a lot to the huge body of linguistic work on verb aspect and event structure \cite{Dowty_1986_effects_of_aspectual_class_on_temporal_structure_of_discourse_semantics_or_pragmatics,Pustejovsky_1991_syntax_of_event_structure,MoensSteedman_1988_Temporal_ontology_and_temporal_reference,Smith_1997_parameter_of_aspect}, verb classes \cite{Vendler_1957_Verbs_and_Times,Levin_1993_English_verb_classes_and_alternations_a_preliminary_investigation,VerbNet}, and particularly the geometric event phase representations by \newcite{Croft_2012_Verbs_aspect_and_causal_structure}. To the best of our knowledge, this is the first attempt to merge aspectual and event order\footnote{\newcite{ReimersDehghaniEtAl_2016_Temporal_Anchoring_of_Events_for_the_TimeBank_Corpus} proposed distinguishing between ``single-day" and ``multi-day" events, but this was to enable anchoring to temporal expressions rather than to annotate event order.} information in a single annotation unit (in TimeML they are separate). %

To achieve this, \name\  distinguishes between bounded, unbounded and partially bounded\footnote{We hope that the linguist reader will excuse our re-defining ``boundedness'', an established term in Aktionsart literature.} events, defined as follows.

\begin{wrapfigure}[]{L}{0.40\textwidth}
    \centering    \includegraphics[width=5.5cm]{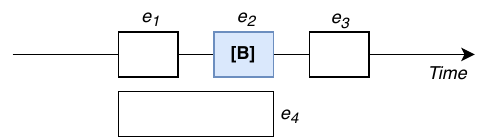}
    \caption{Bounded events}
    \label{fig:b}
\justifying \noindent   
Example: \textit{John started \uline{working}\textsubscript{4} when Mary \uline{came in}\textsubscript{1}, and stopped when she \uline{packed}\textsubscript{2} and \uline{left}\textsubscript{3} for New York}.    
\end{wrapfigure}

\paragraph{Bounded events [B]} are events (of any nature and duration) that are known to start roughly after the end of the nearest other event on the timeline, and they end before the next event starts (with or without a temporal gap). 

In the example shown in \autoref{fig:b}, the event of Mary packing ($e_2$) is ``bounded" by the events of her coming ($e_1$) and leaving ($e_3$). John working is also a bounded event, the duration of which spans $e_1$:$e_2$. The start of $e_1$ and the end of $e_3$ are ``bounded'' by the start/end of the story.

\begin{wrapfigure}[]{L}{0.40\textwidth}
    \centering
    \includegraphics[width=5cm]{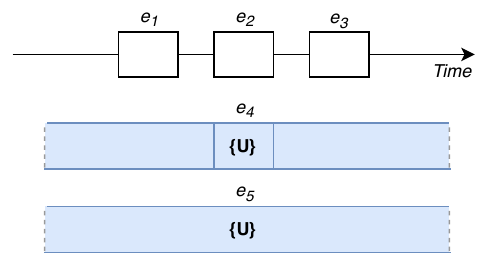}
    \caption{Unbounded events}
    \label{fig:u}
\justifying \noindent    
Example: \textit{Mary \uline{went}\textsubscript{1} to the coffee shop and \uline{found}\textsubscript{2} John there. He was \uline{working}\textsubscript{4} on his lifelong \uline{project}\textsubscript{5}. She \uline{left}\textsubscript{3}.} 
\end{wrapfigure}

\paragraph{Unbounded events \{U\}} are events (of any nature and duration), of which the exact start and end points are not known, but they are known to overlap with some other event on the timeline, and (in an underspecified way) with its nearest neighbors.

In the example in \autoref{fig:u}, the event of John working ($e_4$) started at an underspecified point, possibly before Mary started walking to the coffee shop ($e_1$). We also don't know when he stops working; maybe immediately after Mary's leaving ($e_3$), and maybe hours later. The only thing we know for sure is that he was working when Mary saw him ($e_2$), and this is what \{U\} events encode in \name . The temporal location of [B] event $e_2$ is used as the temporal ``center" of the \{U\} event $e_4$. 

A big advantage of this definition of unbounded events is that it singles out the cases where the exact temporal order is underspecified, but some inference about relations of events surrounding the anchor [B] event and the \{U\} event may be possible based on the world knowledge. Our intuition is that it didn't take Mary long to get to the coffee shop, so John was probably working while she was getting there. Specifying such guesses is not in the scope of event order annotation, but there are relevant efforts collecting data about possible event durations \cite{VashishthaVanDurmeEtAl_2019_Fine-Grained_Temporal_Relation_Extraction} and commonsense reasoning \cite{QinGuptaEtAl_2021_TIMEDIAL_Temporal_Commonsense_Reasoning_in_Dialog,ZhouKhashabiEtAl_2019_Going_on_vacation_takes_longer_than_Going_for_walk_Study_of_Temporal_Commonsense_Understanding}. In the future, we might be able to make better guesses about event durations, and \name\  annotation could tell where such reasoning would be warranted. %
See also the work of  \newcite{LeeuwenbergMoens_2020_Towards_Extracting_Absolute_Event_Timelines_From_English_Clinical_Reports}, who take the opposite approach and directly elicit from the annotators the upper and lower bounds of the target events.

We also define a special case of ``permanent'' unbounded events, represented in this example by event $e_5$ (John's lifelong project). This is an event that occurs throughout the narrative, and likely also beyond it. Such events are also of \{U\} type, but they are not ``centered" on any particular slot on the timeline. We use this mechanism to account for relatively permanent characteristics of characters and entities, which are unlikely to change in the course of the narrative. %
We also use this mechanism for generic events such as ``people like coffee", as they can be conceptualized as occurring all the time.

\begin{wrapfigure}[]{L}{0.40\textwidth}
    \centering    \includegraphics[width=5.5cm]{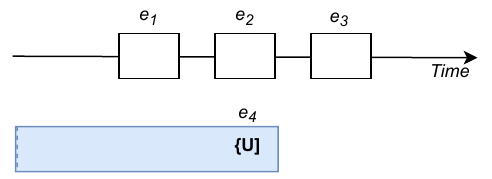}
    \caption{Partially bounded events}
    \label{fig:us}
\justifying \noindent 
Example: \textit{Mary \uline{walked}\textsubscript{1} across the garden. She \uline{called}\textsubscript{2} for John. He stopped \uline{working}\textsubscript{4}, and they \uline{left}\textsubscript{3} together.}
\end{wrapfigure}

\paragraph{Partially bounded events [U\}, \{U]} are a combination of the two above types, used when one endpoint of an event is known, and the other endpoint is underspecified. \autoref{fig:us} illustrates an event bounded on its right endpoint, and unbounded on the left. The event of Mary calling John ($e_2$) is ``anchoring'' the \{U] type event of John's working $e_4$, which lasts during her calling him and for some underspecified time prior to that. He was probably working while she was walking, but that is in the sphere of inference based on world knowledge.

These 3 event types account for the ambiguities between events that can be placed on a coherent timeline. Vagueness due to the different timelines is handled by branches  (\autoref{sec:branches}). 

\name\  annotators are free to choose the level of granularity of event order. %
For example, we might interpret John stopping to work as something that happens \textit{after} Mary calling him: e.g. if we know that John is not someone to spring up instantly, or if it is a crime story where the exact order matters. But the interval is so small that in most cases these events could be considered roughly simultaneous. \name\  framework can accommodate either interpretation, depending on annotator instructions or the saliency of the order between fast-occurring events in the text.

\subsection{Timeline branches}
\label{sec:branches}

\begin{wrapfigure}[]{L}{0.55\textwidth}
    \centering
    \includegraphics[width=7cm]{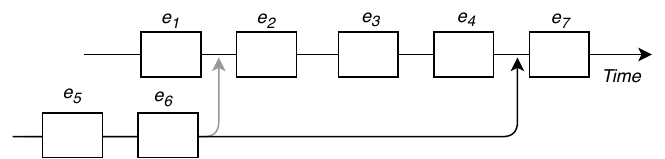}
    \caption{Branching timelines in \name }
    \label{fig:branch}
\justifying \noindent
Example: \textit{John \uline{came}\textsubscript{~$e_1$}  back  to Boston. (...) He \uline{bought}\textsubscript{~$e_2$} a ticket, \uline{had}\textsubscript{~$e_3$} a coffee and \uline{headed}\textsubscript{~$e_4$} to the cinema. He had already \uline{read}\textsubscript{~$e_5$} the book and he \uline{liked}\textsubscript{~$e_6$} it. The movie \uline{started}\textsubscript{~$e_7$}.}
\end{wrapfigure}

The kinds of vagueness about temporal relations that are encoded in bounded/unbounded event span definitions (\autoref{sec:etypes}) can only help with events that are on the same coherent timeline. However, often there is not enough information to build such a timeline, even if the events do not differ by factuality (see \autoref{sec:factuality}). %
Consider the example shown in \autoref{fig:branch}. It is not clear whether John read the book before or after coming to Boston, although we do know that he did it before watching the movie.

\normalsize

\name\  handles such cases by creating a branch on the main timeline. %
A branch is defined as a mini-timeline, linked with a before/after relation to some location on the main timeline. In this example, one such candidate attachment points is the movie visit. The events on the branch happen in parallel to the events in the corresponding section of the main timeline, and are in a \textsc{Vague} relation to them. 

Naturally, we know that it takes longer to read a book than to get to a movie theater, and we could infer that the book was read before the whole movie-related sequence. Whether to perform this extra reasoning step turned out to be a big source of disagreement. We experimented with forcing the annotators to attach branches simply where they were mentioned, but this extra reasoning is a part of natural reading process, and turned out to be hard to suppress consistently. We believe this is one of the reasons why temporal annotation generally suffers from relatively low IAA.

\name\ allows for three types of branches: for event(s) happening at some time before a given point, after a given point, or between two given points on the main timeline.

\subsection{Factuality}
\label{sec:factuality}

So far we have considered events, the temporal order between which can be fully or partially specified, but it is clear that all these events did in fact happen. Another source of uncertainty in the temporal annotation is events for which that is not clear, such as future events, negated events, conditionals, modals, comparisons, and figures of speech.  \newcite{NingWuEtAl_2018_A_Multi-Axis_Annotation_Scheme_for_Event_Temporal_Relations} address that problem by proposing to place events with different realis status on different timelines, so as to avoid annotating underdefined relations.

Our solution is based on the possible-worlds approach: all such events are treated as real events on the timeline for the purposes of establishing temporal order. For example, if a text mentions that John didn't send a birthday present to his mother, this non-event is in fact an event with a certain timeline location. To account for the realis status, we introduce a simplified version of FactBank \cite{SauriPustejovsky_2009_FactBank_Corpus_Annotated_with_Event_Factuality} factuality markup, which  combines the axes of negation (happened/didn't happen) and certainty (did happen/maybe happened). 

\begin{figure*}[!htbp]
    \centering
    \includegraphics[scale=0.39]{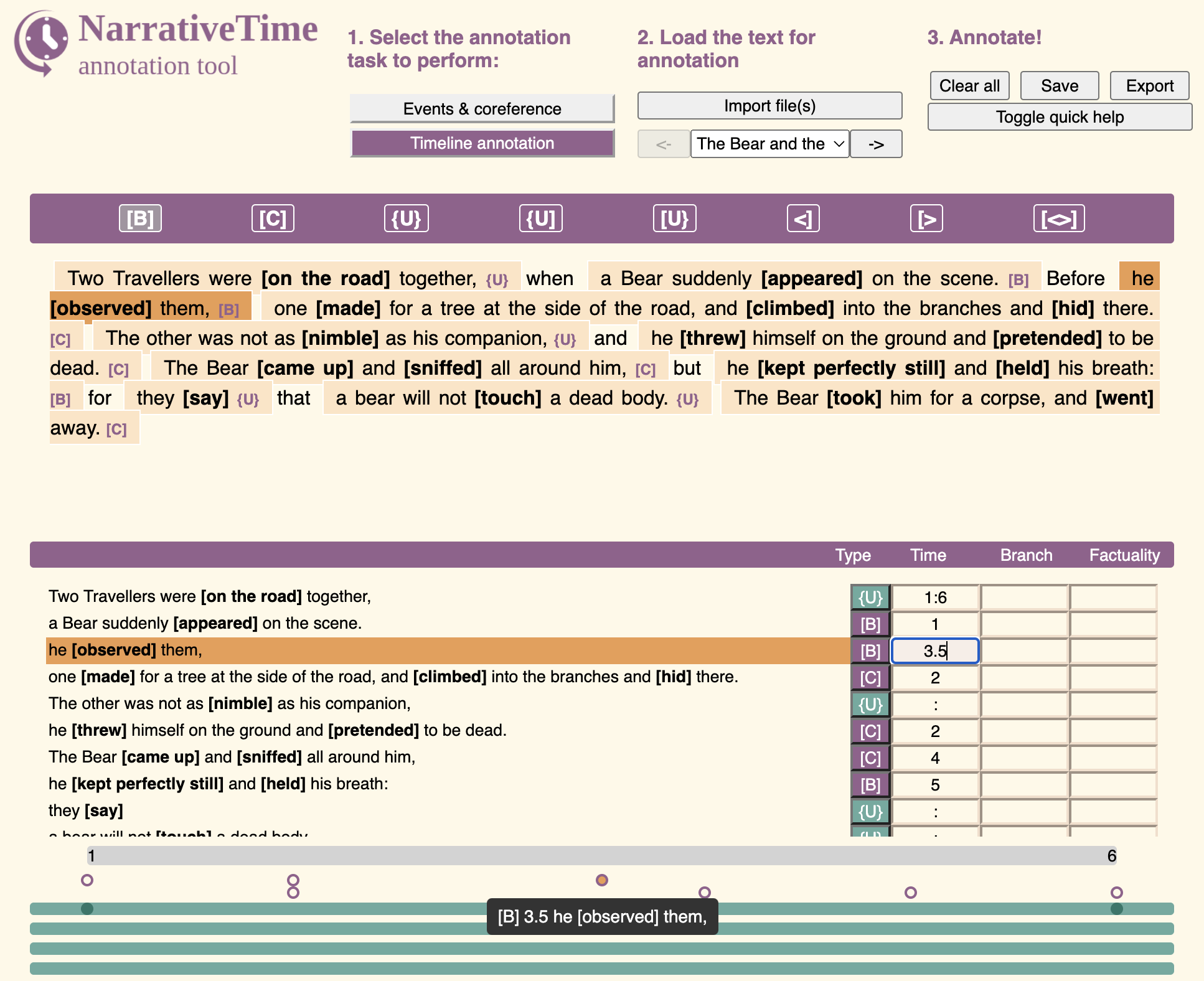}
    \caption{\name\ annotation interface %
    }
    \label{fig:interface}
\end{figure*}

This gives us four possible values for factuality. Since most events in narrative texts are of the ``happened" type, in \name\  they are left unmarked for factuality. The other types can be manually specified in the ``factuality'' column in the annotation interface (\autoref{fig:interface}) 
with the following simple text markers: \textbf{``-"} for ``didn't/won't happen'', \textbf{``m"} for ``maybe happened/will happen", and \textbf{``m-"} for ``maybe didn't/won't happen". %

\subsection{Event clusters}
\label{sec:clusters}

Psychologists established that texts that are pre-chunked in semantically coherent segments are easier to process \cite{FraseSchwartz_1979_Typographical_cues_that_facilitate_comprehension,OSheaSindelar_1983_The_Effects_of_Segmenting_Written_Discourse_on_the_Reading_Comprehension_of_Low-_and_High-Performance_Readers,RajendranDuchowskiEtAl_2013_Effects_of_text_chunking_on_subtitling_A_quantitative_and_qualitative_examinationa}. %
For dynamic situations in the narratives, we hypothesize that ``semantic coherence'' is best explained in terms of scripts/frames. %
For example, the sentence ``John woke up, brushed his teeth, got dressed, went to the office, and proposed to Mary", is likely to be remembered as 2 events rather than 5: the morning-routine event and the proposal event.%

\name\  leverages this feature of human reading comprehension by encouraging the annotators to think in terms of event clusters rather than single events. In particular, we define the following types of event clusters that can be used in annotation:

\begin{itemize*}
    \item \textbf{Clusters of roughly-simultaneous bounded events.} A [B] event can denote either a single bounded event, or a cluster where the events are either roughly-simultaneous, or their order does not matter for the purposes of the current narrative. For example, in the sentence  \textit{John \uline{called}, \uline{texted} and \uline{left voicemails} for Mary incessantly} the order of these actions does not matter.  %
 
    \item \textbf{Clusters of consecutive events.} Narratives often contain mini-scripts (\textit{John \uline{brushed} his teeth and \uline{got dressed}}), or combinations of cause/effect, enabling/enabled events that could only happen in that order (\textit{John \uline{woke up} and \uline{thought of} Mary.}) We defined a special event type [C] which amounts to a sequence of \uline{c}onsecutive [B]... [B] events.
 
    \item \textbf{Clusters of unbounded events.} Narratives often contain descriptive sequences, such as "\textit{John was a \uline{short}, \uline{fat} man with a \uline{red face} and a \uline{bald patch}}". The temporal information for all these features is the same, so they can all be annotated as a single \{U\} event.
\end{itemize*}

Whether a particular sequence would be processed as a cluster or a sequence of individual events is up to the annotator, and could be expected to differ based on their cognitive styles. But the annotators who chunk the text differently could still produce annotations that are equivalent in terms of event order sequence on the timeline. %

\subsection{Annotation workflow}
\label{sec:annotation}

\name\  comes with a new open-source web-based tool for annotation, available in the project repo.\footnote{\url{https://github.com/text-machine-lab/nt}} %
The interface for annotating event order (with events already pre-marked) is shown in \autoref{fig:interface}.  There are four main elements: 
\begin{itemize*}
    \item The event type choice panel: \textbf{[B]} for bounded events, \textbf{[C]} for clusters of consecutive bounded events, \textbf{\{U\}}, \textbf{[U\}} and \textbf{\{U]} for fully and partially unbounded events, and \textbf{$\langle$]}, \textbf{[$\rangle$}, \textbf{[$\langle\rangle$]} for the three types of branches.  
    \item The text area, which shows pre-marked \textbf{[events]} and highlighted text spans corresponding to event clusters.
    \item The annotation table, which lists the text of all annotated spans and their values for timeline positions (\textit{time} column), branch anchors (\textit{branch} column), and factuality values (\textit{factuality} column). All of these can be manually edited.
    \item The interactive timeline representation of all existing annotations, with bounded events shown as purple elements and unbounded - as green elements. The text of spans corresponding to any timeline element can be viewed in a tooltip by hovering over that element, as shown in the screenshot.
\end{itemize*}

An annotation is created by choosing the event type ([B] by default), highlighting some span in the text, and either accepting the auto-populated values of time, branch, and factuality, or manually editing them in the annotation table. By default, any new annotation is a bounded, actual event on the main timeline, at the position following the previous highest one (e.g. if the timeline currently ends at position 2, then a new [B] event will be placed at 3).

This workflow minimizes the number of clicks required from the annotators: the best case scenario is that they only need to read the text, highlighting events in chronological order. That will auto-populate the timeline integers serving as timeline position indicators. To ``move'' an event to another timeline position only its \textit{time} value needs to be edited. This way it is easy to insert new events without changing existing annotations: e.g. if there are events at positions 1 and 2, a new event can be placed between them by setting its \textit{time} value to 1.5. The type of an existing annotation can also be changed (by clicking on the type button in the annotation table).

It is possible to annotate the order of individual events, only highlighting their spans, but, as shown in the example in \autoref{fig:interface}, the tool also allows annotating spans of text containing several events, and such spans are interpreted as clusters of bounded, unbounded, or bounded consecutive events (\autoref{sec:clusters}). This both saves annotation effort, and allows to leverage the natural chunking-during-reading  strategies of the annotators.

In this study, we used pre-annotated events and event coreference information from the original TimeBank, but our annotation tool also has a basic interface for the first step of the annotation process: 
selecting events, the order of which needs to be annotated, and specifying their coreference information. 

\subsection{Anchoring of temporal expressions}
\label{sec:timex}

\name\  follows  \newcite{PustejovskyIngriaEtAl_2005_specification_language_TimeML} in defining temporal expressions (timex). We make no contribution in this area, and use the pre-existing timex annotations of TimeBank in our case study. What \name\  does improve is their linking with events: annotators only need to include any temporal expressions in the event spans which they anchor, so the spans function as temporal containers \cite{PustejovskyStubbs_2011_Increasing_Informativeness_in_Temporal_Annotationa}. No further action is needed for event-timex links. 

For example, if [\textit{John \uline{met} Mary on \textbf{Monday}}] is chosen as the event span, then the meeting event would be anchored to Monday. If a cluster of simultaneous events is in the same span as a timex, then all of them are anchored to that timex.%
This approach echoes treating temporal expressions as event arguments, which reportedly reduces the annotation effort by 85\% as compared to TimeBank-Dense \cite{ReimersDehghaniEtAl_2016_Temporal_Anchoring_of_Events_for_the_TimeBank_Corpus}. %
If a timex applies to several consecutive events (e.g. from timeline position 2 to 5), it is possible to create a separate span for that timex and specify its duration as an interval (e.g. 2:5).

\subsection{Post-processing}
\label{sec:post-processing}

Given our new definitions of event types, we developed a new representation for \name\  annotation that is used internally in the annotation tool. This is a simple json-based format containing the indices of pre-annotated timexes, events, and their coreference chains, as well as the indices and timeline positions, types, actuality, and branch annotations for the timeline annotations. A small example of this format is shown in \autoref{nt_format}; 
see the project repository for more details.

The internal format allows for underspecification in temporal relations through the \name\ mechanisms (branches, factuality, and unbounded events). However, the current standard for representing temporal information is based on event-event or event-time pairs, specifically, TimeML-ISO \cite{PustejovskyLeeEtAl_2010_ISO-TimeML_An_International_Standard_for_Semantic_Annotation}, and this is what most existing applications expect. Hence we also provide a tool for converting the \name\ annotation  to the more familiar TimeML \tlinks\ (see the project repository for details). We opted to use 5 classic TimeML relations (\textsc{before}/\textsc{after}, \textsc{includes}/\textsc{is\_included}, \textsc{simultaneous}), as well as \textsc{vague} \cite{VerhagenGaizauskasEtAl_2007_SemEval-2007_Task_15_TempEval_Temporal_Relation_Identification} and \textsc{overlap} \cite{VerhagenGaizauskasEtAl_2007_SemEval-2007_Task_15_TempEval_Temporal_Relation_Identification}
Without the inverse relations (\textsc{before}/\textsc{after}, \textsc{includes}/\textsc{is\_included}), the set could be reduced to 5. This mapping is external and auxiliary to \name, and other mappings could also be developed.

\autoref{tml_format} shows the data from \autoref{nt_format} represented in with TimeML (for text and \textsc{TLink} tags) and FactBank (for FACT\_VALUE tags) style. This is a small example with only 4 events and 1 timex, and we do not show the possible inverse relations (which would double the overall amount of \tlinks), but the explicit enumeration of all possible \tlinks\ still looks more verbose, and harder to fix errors in.

The format conversion also involves significant conceptual trade-offs, since it requires a mapping between \name\  format, which represents the vague relations with the combination of unbounded events and branching mechanism, and the classical TimeML relations. Our choices are shown in \autoref{tab:conversion}, with examples of overlapping and non-overlapping temporal intervals indicating the timeline positions for different combinations of event types. 

The first column (the case of two bounded events \textbf{[][]}) is simple and corresponds to the classical TimeML relations, but the cases involving unbounded events (\textbf{[\}, \{]} and \textbf{\{\}}) are more difficult. We opted to map to \textsc{vague} (empty cell in the table) all cases where more than one relation could theoretically be possible: for example, an unbounded event at position \{3\} necessarily \textsc{Includes} a bounded event at position [3], but its position with respect to another unbounded event at position \{3\} could be either \textsc{simultaneous} or \textsc{overlap}, depending on the exact edges of the two events (underspecified by definition, could only be resolved with case-by-case commonsense reasoning or by providing more contextual information). 

\begin{Code}
    \centering
\begin{lstlisting}[language=Python,label=nt_format, caption=NarrativeTime native format example]
{
    # text id
    "id":"sample",

    # space-tokenized text
    "text":"John ordered a new bike for his summer trip , but his order got lost .",

    # "spring" timex annotation: [start token, end token] 
    "timex": {"0": [7, 7]},

    # similarly structured event annotations ("ordered", "used", "order", "lost")
    "events":{"0":[1,1],"1":[8,8],"2":[12,12],"3":[14,14]}, 

    # coreference chain between "ordered" (token 1) and "order" (token 12)
    "event_coreference":{"1":[12]},

    # events in coreference chains are unmarked for annotation, except for the first mention
    "invisible_events": [12],

    # timeline annotation
    "event_order":{
        "0":{"span":[0,4],"type":0,"time":"1","factuality":"","branch":""},
        "1":{"span":[6,8],"type":0,"time":"3","factuality":"m","branch":""},
        "2":{"span":[11,14],"type":0,"time":"2","factuality":"","branch":""}}
    # "span": [start token, end token] for the annotated span
    # "type": the span types (0=[B], 1=[C], 3={U}, 4=[U}, 5={U])
    # "time": the timeline position of the annotated span	
    # "factuality": factuality annotation
    # "branch": the timeline attachment point of a branch + its type
}
\end{lstlisting}

\begin{lstlisting}[language=xml,label=tml_format,  caption=Listing 1 data represented in TimeML and FactBank style]
<?xml version="1.0" encoding="utf-8"?>

<TimeML>

John <EVENT eid="0">ordered</EVENT>a new bike for his <TIMEX3 tid="t0">summer</TIMEX3><EVENT eid="1">trip</EVENT>, but his <EVENT eid="2">order</EVENT>got <EVENT eid="3">lost</EVENT> . 

<MAKEINSTANCE eiid="ei0" eventID="0"/>
<MAKEINSTANCE eiid="ei1" eventID="1"/>
<MAKEINSTANCE eiid="ei2" eventID="2"/>
<MAKEINSTANCE eiid="ei3" eventID="3"/>

<FACT_VALUE eiid="ei0" fvid="1" value="CT+"/>
<FACT_VALUE eiid="ei1" fvid="2" value="PS+"/>
<FACT_VALUE eiid="ei3" fvid="3" value="CT+"/>
<FACT_VALUE eiid="ei2" fvid="4" value="CT+"/>

<TLINK lid="1" eventInstanceID="ei0" relType="BEFORE" relatedToEventInstance="ei1"/>
<TLINK lid="2" eventInstanceID="ei0" relType="BEFORE" relatedToTime="t0"/>
<TLINK lid="3" eventInstanceID="ei0" relType="BEFORE" relatedToEventInstance="ei3"/>
<TLINK lid="4" eventInstanceID="ei0" relType="SIMULTANEOUS" relatedToEventInstance="ei2"/>
<TLINK lid="5" eventInstanceID="ei1" relType="AFTER" relatedToEventInstance="ei0"/>
<TLINK lid="6" eventInstanceID="ei1" relType="SIMULTANEOUS" relatedToTime="t0"/>
<TLINK lid="7" eventInstanceID="ei1" relType="AFTER" relatedToEventInstance="ei3"/>
<TLINK lid="8" eventInstanceID="ei1" relType="AFTER" relatedToEventInstance="ei2"/>
<TLINK lid="9" timeID="t0" relType="AFTER" relatedToEventInstance="ei0"/>
<TLINK lid="10" timeID="t0" relType="SIMULTANEOUS" relatedToEventInstance="ei1"/>
<TLINK lid="11" timeID="t0" relType="AFTER" relatedToEventInstance="ei3"/>
<TLINK lid="12" timeID="t0" relType="AFTER" relatedToEventInstance="ei2"/>
<TLINK lid="13" eventInstanceID="ei3" relType="AFTER" relatedToEventInstance="ei0"/>
<TLINK lid="14" eventInstanceID="ei3" relType="BEFORE" relatedToEventInstance="ei1"/>
<TLINK lid="15" eventInstanceID="ei3" relType="BEFORE" relatedToTime="t0"/>
<TLINK lid="16" eventInstanceID="ei3" relType="AFTER" relatedToEventInstance="ei2"/>
<TLINK lid="17" eventInstanceID="ei2" relType="SIMULTANEOUS" relatedToEventInstance="ei0"/>
<TLINK lid="18" eventInstanceID="ei2" relType="BEFORE" relatedToEventInstance="ei1"/>
<TLINK lid="19" eventInstanceID="ei2" relType="BEFORE" relatedToTime="t0"/>
<TLINK lid="20" eventInstanceID="ei2" relType="BEFORE" relatedToEventInstance="ei3"/>

</TimeML>
\end{lstlisting}
\end{Code}
\begin{table}[!t]
\adjustbox{max width=\textwidth}{%
\def\arraystretch{1.2}
\begin{tabular}{>{\columncolor[HTML]{EFEFEF}}p{1.3cm}>{\columncolor[HTML]{EFEFEF}}p{1.3cm}p{2.4cm}p{2.4cm}p{2.4cm}p{2.4cm}p{2.4cm}p{2.4cm}}
\toprule
\textsc{$e_1$ time} & \textsc{$e_2$ time}  & \textbf{\textsc{{[}$e_1${]} {[}$e_2${]}}} & \textbf{\textsc{\{$e_1$\} \{$e_2$\}}} & \textbf{\textsc{{[}$e_1$\} {[}$e_2$\}}} & \textbf{\textsc{\{$e_1${]} \{$e_2${]}}} & \textbf{\textsc{{[}$e_1${]} \{$e_2$\}}} & \textbf{\textsc{\{$e_1$\} {[}$e_2${]}} }\\
\midrule
1:3 & 4:6 & \textsc{before} &  &  &  &  &  \\
4:6 & 1:3 & \textsc{after} &  &  &  &  &  \\
1:6 & 3:4 & \textsc{\textsc{includes}} &  &  &  &  & \textsc{includes} \\
3:4 & 1:6 & \textsc{is\_included} &  &  &  & \textsc{is\_included} &  \\
1:4 & 3:6 & \textsc{overlap} &  &  &  &  &  \\
3:6 & 1:4 & \textsc{overlap} &  &  &  &  &  \\
1:3 & 1:3 & \textsc{simultaneous} &  &  &  & \textsc{is\_included} & \textsc{includes}
\end{tabular}
}

\vspace{.5cm}

\adjustbox{max width=\textwidth}{%
\def\arraystretch{1.2}
\begin{tabular}{>{\columncolor[HTML]{EFEFEF}}p{1.3cm}>{\columncolor[HTML]{EFEFEF}}p{1.3cm}p{2.4cm}p{2.4cm}p{2.4cm}p{2.4cm}p{2.4cm}p{2.4cm}}
\toprule
\textsc{$e_1$ time} & \textsc{$e_2$ time} & \textbf{\textsc{{[}$e_1${]} {[}$e_2${]}}} & \textbf{\textsc{{[}$e_1${]} {[}$e_2$\}}} & \textbf{\textsc{{[}$e_1$\} {[}$e_2${]}}} & \textbf{\textsc{{[}$e_1${]} \{$e_2${]}}} & \textbf{\textsc{\{$e_1${]} {[}$e_2${]}}} & \textbf{\textsc{\{$e_1$\} {[}$e_2$\}}} \\
\midrule
1:3 & 4:6 & \textsc{before} & \textsc{before} &  &  & \textsc{before} &  \\
4:6 & 1:3 & \textsc{after} &  & \textsc{after} & \textsc{after} &  &  \\
1:6 & 3:4 & \textsc{includes} &  & \textsc{includes} &  & \textsc{includes} &  \\
3:4 & 1:6 & \textsc{is\_included} & \textsc{is\_included} &  & \textsc{is\_included} &  &  \\
1:4 & 3:6 & \textsc{overlap} & \textsc{overlap} &  &  & \textsc{overlap} &  \\
3:6 & 1:4 & \textsc{overlap} &  & \textsc{overlap} & \textsc{overlap} &  &  \\
1:3 & 1:3 & \textsc{simultaneous} & \textsc{is\_included} & \textsc{includes} & \textsc{is\_included} & \textsc{includes} & \textsc{includes}
\end{tabular}
}

\vspace{.5cm}

\adjustbox{max width=\textwidth}{%
\def\arraystretch{1.2}
\begin{tabular}{>{\columncolor[HTML]{EFEFEF}}p{1.3cm}>{\columncolor[HTML]{EFEFEF}}p{1.3cm}p{2.4cm}p{2.4cm}p{2.4cm}p{2.4cm}p{2.4cm}p{2.4cm}}
\toprule
\textsc{$e_1$ time} & \textsc{$e_2$ time} & \textbf{\textsc{{[}$e_1${]} {[}$e_2${]}}} & \textbf{\textsc{{[}$e_1$\} \{$e_2$\}}} & \textbf{\textsc{\{$e_1$\} \{$e_2${]}}} & \textbf{\textsc{\{$e_1${]} \{$e_2$\}}} & \textbf{\textsc{\{$e_1${]} {[}$e_2$\}}} & \textbf{\textsc{{[}$e_1$\} \{$e_2${]}}} \\
\midrule
1:3 & 4:6 & \textsc{before} &  &  &  & \textsc{before} &  \\
4:6 & 1:3 & \textsc{after} &  &  &  &  & \textsc{after} \\
1:6 & 3:4 & \textsc{includes} &  &  &  &  &  \\
3:4 & 1:6 & \textsc{is\_included} &  &  &  &  &  \\
1:4 & 3:6 & \textsc{overlap} &  &  &  & \textsc{overlap} &  \\
3:6 & 1:4 & \textsc{overlap} &  &  &  &  & \textsc{overlap} \\
1:3 & 1:3 & \textsc{simultaneous} & \textsc{is\_included} & \textsc{includes} & \textsc{is\_included} & \textsc{overlap} & \textsc{overlap}
\end{tabular}
}
\caption{Mapping of \name\ interval relations to TimeML relations. The first two columns show examples of overlapping and non-overlapping temporal intervals indicating timeline positions of events (a single-value position X is equivalent to X:X interval, e.g. 3:3.) %
The remaining columns show different combinations of event types with these intervals. Empty cells indicate the \textsc{vague} relation. }
\label{tab:conversion}
\end{table}

As evident from \autoref{tab:conversion}, this means losing information, since \name\ format can express the difference between the vagueness on both or one end\footnote{In the pairwise approach, the partial unboundedness could be partially implemented by introducing additional \textsc{start\_on} and \textsc{end\_by} relations, but this would require an additional \tlink\ to specify the \textsc{vague} relation at the other end of the interval. If such an event is ``centered'' on several other events rather than one, even more annotation would be needed.} of an unbounded event. It also does not allow for differentiation between vagueness due to unboundedness and branching. Future work could explore learning/predicting temporal information directly from \name\ representation, or developing more fine-grained types of \textsc{vague} for the classical TimeML representation.

For the events in the branches, their relations with events/timexes on the main timeline is determined by their anchor position and their direction. For example, if a branch is anchored at position 3 and goes into the future, its events are \textsc{after} any main timeline events prior to 3, and \textsc{vague} with the events after position 3 (since they exist in a parallel world, so to speak).
\section{\name\ annotation}

\subsection{TimeBankNT corpus}
\label{sec:corpus}

In scope of this work, we re-annotate 36 documents of the TimeBank corpus which were also used in TimeBank-Dense \cite{CassidyMcDowellEtAl_2014_An_Annotation_Framework_for_Dense_Event_Ordering}, MATRES \cite{NingWuEtAl_2018_A_Multi-Axis_Annotation_Scheme_for_Event_Temporal_Relations} and TDDiscourse  \cite{NaikBreitfellerEtAl_2019_TDDiscourse_Dataset_for_DiscourseLevel_Temporal_Ordering_of_Events}. This enables direct comparison between the number of annotated \tlinks\ with the different methodologies.

\begin{wrapfigure}{r}{0.4\textwidth}
\begin{center}
\begin{tabular}{l|r}
    \toprule
    Texts & 36 \\
    Events & 1,715 \\
    Timexes & 289 \\
    \midrule
    Event-event \tlinks\ & 79,001 \\
    Event-timex \tlinks\ & 23,979 \\
    Timex-timex \tlinks\ & 1,770 \\
    \midrule
    Factuality annotations & 1,715 \\
    \bottomrule
\end{tabular}
\captionof{table}{TimeBankNT corpus statistics}
\label{tab:stats}
\end{center}
\end{wrapfigure}
\FloatBarrier

Two first authors of this paper were the annotators. The guidelines were developed iteratively and underwent many rounds of revision,
which involved annotating the same documents and discussing the cases of disagreement, both in news and fiction texts. 

After that, the full set of 36 TimeBank-Dense documents were annotated independently by the two annotators. This approach enables more reliable estimates of inter-annotator agreement, as well as its variation by annotator and by text. Furthermore, given that many disagreements are  genuine (not due to errors), as will be shown in \autoref{sec:qualitative}, such data is important for the new generation of NLP tools that embrace human label variation \cite{UmaFornaciariEtAl_2021_Learning_from_Disagreement_Survey,Plank_2022_Problem_of_Human_Label_Variation_On_Ground_Truth_in_Data_Modeling_and_Evaluation}.

The final annotation statistics for the corpus are shown in Table \ref{tab:stats}. See \autoref{fig:relation_type_distribution} for the distribution of \tlinks\ labels.

\subsection{Inter-annotator agreement}
\label{sec:iaa}

We compute four types of inter-annotator agreement: event type, factuality, branching and event order.
For event types, we compare if both annotators chose the same type  (e.g., [U\}) for the given event.  
For event order, we convert\footnote{Since the clustering mechanism of \name\ allows for different span annotations with equivalent timelines (\autoref{sec:clusters}), computing agreement directly on span annotations would  reflect not only the temporal order, but also the individual differences in chunking strategies.} \name\ annotation to TimeML using the approach described in \autoref{sec:post-processing}, and compare all event-event and event-timex \tlinks\ for all 7 relation types in our conversion scheme (\autoref{sec:post-processing}). This tests both timeline and event type annotation, as event relations depend on both. 
For factuality we compare whether the given event has the same factuality annotation (including the default empty value, which corresponds to non-negated actual events). For branching we compare if both annotators decided to place event to a branch instead of the main timeline.
The results are shown in \autoref{tab:agreement_results}. 

\begin{table}[ht]
\centering
    \begin{tabular}{l|c|c|c|c}
    \toprule
    & \textsc{Event Type} & \textsc{Event Order} & \textsc{Factuality} & \textsc{Branching}\\
    \midrule
    Agreement Rate & 0.88 & 0.75 & 0.93 & 0.92 \\
    Cohen's $\kappa$ & 0.62 & 0.68 & 0.84 & 0.68 \\
    Krippendorff's $\alpha$\tablefootnote{Both agreement coefficients reported here, $\kappa$ and $\alpha$, are values computed on the entire dataset, not averages of values for each document.} & 0.62 & 0.68 & 0.83 & 0.68\tablefootnote{The binary nature of branching (the decision whether or not to place an event on a branch) makes the data distribution naturally skewed as the majority of events are on the main timeline. Computing agreement coefficient such as $\alpha$ or $\kappa$ on skewed distribution results here in lower agreement as represented by these coefficients, which ultimately creates the relatively big gap between the agreement rate (0.98) and agreement coefficients (0.68) \cite{di-eugenio-glass-2004-squibs,Paun2022-bi}.} \\
    \bottomrule
    \end{tabular}
    \caption{Inter-annotator agreement in \name.}
    \label{tab:agreement_results}
\end{table}

\begin{figure}
\centering
\begin{minipage}{.48\textwidth}
\centering
\includegraphics[width=0.75\textwidth]{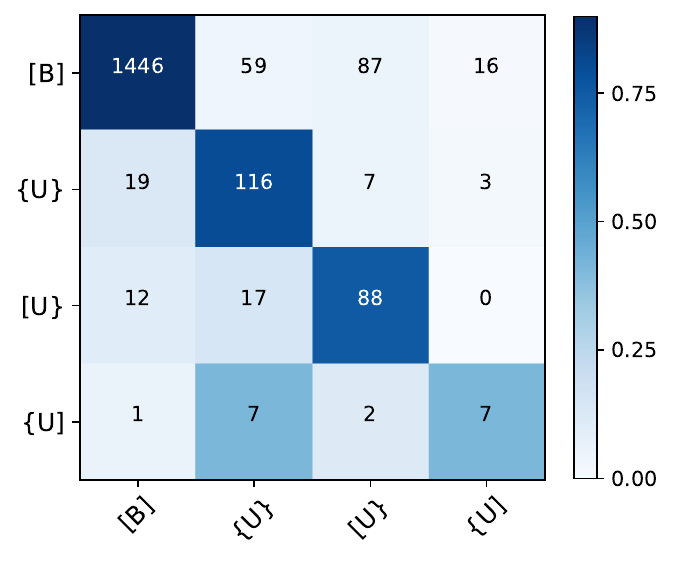}
\captionsetup{skip=35pt}
\caption{\name\ event type confusion matrix for annotation}
\label{fig:event_type_confusion_matrix}
\end{minipage}%
\hfill
\begin{minipage}{.48\textwidth}
\centering
\includegraphics[width=1.0\textwidth]{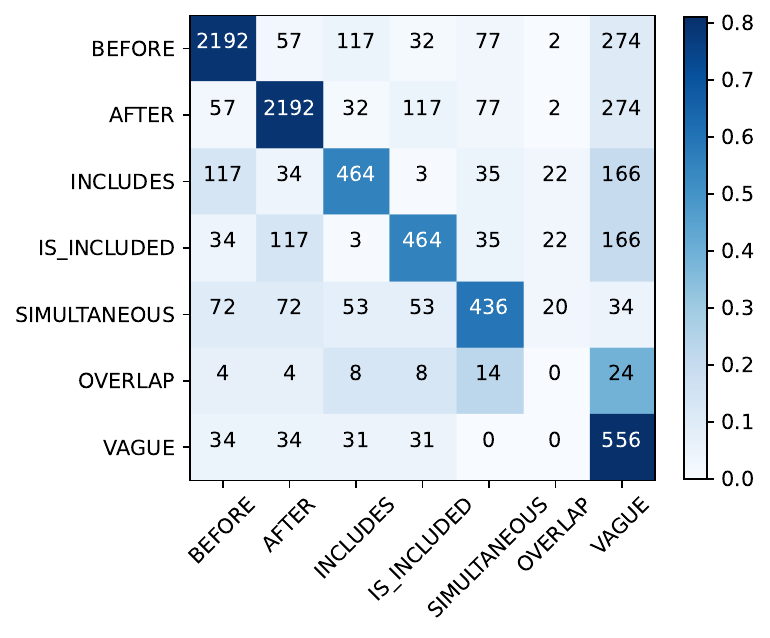}
\caption{\tlink\ relation type confusion matrix for annotation}
\label{fig:tlink_confusion_matrix}
\end{minipage}
\end{figure}

Our results for event type, event order and branching could be described as ``substantial agreement'', and for factuality - as ``perfect agreement'' \cite{Landis1977,artstein-poesio-iaa-2008}. The prior results for temporal order annotation (with IAA estimated as Cohen $\kappa$ or Krippendorff $\alpha$) are in the range of 0.47-0.84 (see \autoref{tab:related_work}). However, the direct comparison with annotation of event pairs is not entirely fair to \name, because we are solving a more difficult task: with a timeline, the annotators have to guarantee that a given annotation is consistent with all other existing annotations, which is not the case in pairwise approach. We also achieve much higher density (\autoref{sec:density}).

The confusion matrices for event types (\autoref{fig:event_type_confusion_matrix}) and \tlink\ relations (\autoref{fig:tlink_confusion_matrix}) should be interpreted in view of the class imbalance in event types and labels: they both show the highest confusion rate for a rare class (\{U] and \textsc{Overlap}, respectively), where even a small number of discrepancies would have a dramatic effect. In fact, the most confusion in event types is between bounded ([B]) and unbounded (\{U\}) or partially bounded ([U\}) events. For the temporal relations, much confusion focuses on \textsc{simultaneous} vs \textsc{includes/Is\_included}and \textsc{simultaneous} vs \textsc{before}/\textsc{after}. The latter is partly explained by a frequent case of genuine (i.e. not due to errors) label variation that we identify in \autoref{sec:qualitative}.

Additionally, we perform a full qualitative evaluation of annotations in 6 documents that vary in IAA values (see \autoref{sec:qualitative}), which will show that many ``disagreements'' are genuine and would be more appropriately described as ``human label variation'' \cite{Plank_2022_Problem_of_Human_Label_Variation_On_Ground_Truth_in_Data_Modeling_and_Evaluation}. This brings up the question of what level of agreement is in principle achievable in full temporal annotation of realistic texts.

\subsection{The use of \name-specific annotation mechanisms}
\label{sec:validation}

As described in \autoref{sec:nt}, \name\ proposes three mechanisms for handling underspecification: unbounded and partially bounded event type, branching, and factuality. We interpret our  substantial agreement on event types and branching, and perfect agrement on factuality (\autoref{tab:agreement_results}), as evidence that the guidelines were sufficiently clear, and the annotators made use of these mechanisms in similar ways. 

\begin{figure}[!b]
    \centering
    \includegraphics[scale=0.39]{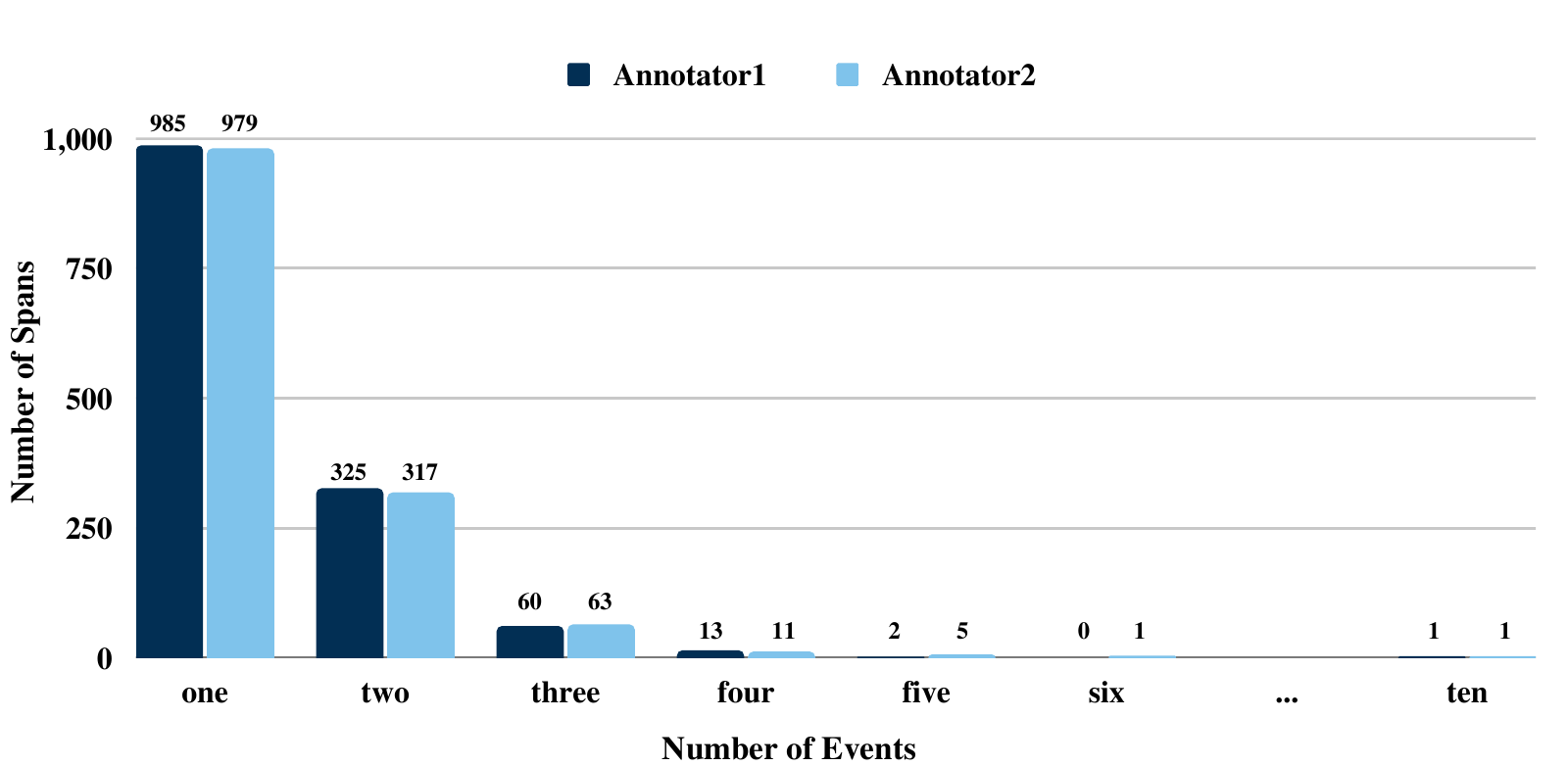}
    \caption{The number of span-based \name\ timeline annotations with the number of events included in the spans. While the majority of spans contained only one event, almost one third encoded two or more events.}
\label{fig:event_span_count}
\end{figure}

A key innovation in \name\ framework is that it enables the annotation of event clusters (\autoref{sec:clusters}), rather than just individual events, which enables annotating multiple temporal relations at once. At the same, time, whether to use this mechanism is up to the annotator, and it is possible to produce equivalent timelines with different chunking strategies.

\autoref{fig:event_span_count} shows the overall distribution of events in the spans highlighted by both annotators: while the majority of annotations contain only one event, almost one third of annotations contain two or more events. The distribution is very similar for the two annotators. We interpret this as evidence that the span-based annotation is useful even in the annotation of news, and we hypothesize that it could be even more useful in other types of text with more temporally coherent chunks of text, such as descriptive paragraphs in fiction or historical narratives in encyclopedias.

\subsection{Annotation density}
\label{sec:density}

A key feature of \name\ is that its timeline representation forms a \textit{complete} temporal representation of the text, explicitly marking any underspecified relations. This has not been possible in the previous approaches based on event pairs, because it would require a number of \tlink\ annotations quadratic to the number of events and timexes in the text. 

\autoref{tab:corpus} shows the base statistics and \tlink-to-event ratio for the densest, to our knowledge, currently available English resources with temporal annotation.  
Among them, the densest expert-annotated resources are TimeBank-Dense \cite{CassidyMcDowellEtAl_2014_An_Annotation_Framework_for_Dense_Event_Ordering} and the recent MAVEN-ERE \cite{maven-ere}. %

\begin{table}[!t]
\setlength{\extrarowheight}{3pt}
\small \centering
\begin{threeparttable}
\begin{tabular}{p{6cm}rrrr}
\toprule
\textsc{Project} & \textsc{Events} & \textsc{Timexes} & \tlinks\ & \textsc{Ratio} \\ \midrule
TempEval-3 \cite{UzZamanLlorensEtAl_2012_Tempeval3_Evaluating_events_time_expressions_and_temporal_relations} & 11,145 & 2,078 & 11,096 & 0.84 \\
UDS-T \cite{VashishthaVanDurmeEtAl_2019_Fine-Grained_Temporal_Relation_Extraction} & 32,302 & -- & 70,368 & 2.20 \\
TimeBank-Dense \cite{CassidyMcDowellEtAl_2014_An_Annotation_Framework_for_Dense_Event_Ordering} & 1,729 & 289 & 12,715 & 7.40 \\
TDDiscourse \cite{NaikBreitfellerEtAl_2019_TDDiscourse_Dataset_for_DiscourseLevel_Temporal_Ordering_of_Events} & ~1,729\tnote{1} & 289 & 6,150 & ~3.05\\
MATRES \cite{NingWuEtAl_2018_A_Multi-Axis_Annotation_Scheme_for_Event_Temporal_Relations} & 6,099%
& 1,955
& 13,577%
& 1.69%
\\ 
TDT-Crd \cite{ZhangXue_2019_Acquiring_Structured_Temporal_Representation_via_Crowdsourcing_Feasibility_Study} & 2,691 & 1,414 & 4,105 & 1.0\\
TDG \cite{YaoQiuEtAl_2020_Annotating_Temporal_Dependency_Graphs_via_Crowdsourcing} &  14,974 & 2,485 & 28,350 & 1.62 \\
Event Storyline \cite{CaselliVossen_2017_The_Event_StoryLine_Corpus_A_New_Benchmark_for_Causal_and_Temporal_Relation_Extraction} & 7,275 & 1,297 & 4,017 & 0.47 \\
MAVEN-ERE \cite{maven-ere} & 103,193 & 25,843 & 1,216,217 & 9.43 \\
\name\  & 1,715\tnote{2} & 289 & 102,313 & \textbf{51.05} \\
\bottomrule
\end{tabular}
\caption{Density of \tlinks\ backed by manual annotation in the current English resources. The density is computed as total number of \tlinks\ (without inverses), divided by  (number of events + number of timexes).
} 
\label{tab:corpus}
\vspace{6pt}
\begin{tablenotes}
\item[1] TDDiscourse paper does not state the number of events; it is probably slightly smaller than in TimeBank-Dense, since their released data reference event IDs rather than event instance IDs. Only event-event \tlinks\ seem to be annotated.
\item[2] The small discrepancy in event number between TimeBankNT and  TimeBankDense is due to the fact that \name\ annotation relies on event tokens rather than event instance tags (although we use the original TimeBank event instance id numbers in conversion).
\end{tablenotes}
\end{threeparttable}
\end{table}

Our solution is 5 times denser than than the previous densest solution, MAVEN-ERE. %
\autoref{tab:corpus} reports only the number of event-event \tlinks\ without inverse relations; with them, the total number of \tlinks\ in TimeBankNT reaches 207,496. %

As discussed in \autoref{sec:review}, the sparsity problem with annotation based on event pairs is usually addressed by trying to infer the missing relations by transitive closure. With such inferred relations, the above-cited resources could be represented as much larger in terms of \tlinks. We believe it would not be a fair comparison: although \name\ also infers \tlinks\ from the timeline, our framework guarantees that the entire timeline is considered by the annotator, and so in effect, all of the inferred \tlinks\ are backed by manual annotation. In temporal closure, they are only backed by the closure rules, and because of incomplete, conflicting, or missing annotations, the full temporal graph often cannot be constructed \cite{OcalPerezEtAl_2022_Holistic_Evaluation_of_Automatic_TimeML_Annotators}.

\subsection{Qualitative Analysis}
\label{sec:qualitative}

We manually analyzed 6 documents (%
4,336 \tlinks)\footnote{The agreement, on \tlinks, on documents sampled for the qualitative analysis, ranges from Krippendorff's $\alpha$=0.47 (one of the lowest) to $\alpha$=0.85 (one of the highest). Choosing documents with varying agreements allows us to analyze not only cases where the annotators' interpretations differ but also cases where they interpret the timeline unanimously.} to identify the cases where annotators' interpretations differ resulting in label variation.\footnote{Here we use the term ``variation'' rather than ``disagreement'' following a recent proposal in \newcite{Plank_2022_Problem_of_Human_Label_Variation_On_Ground_Truth_in_Data_Modeling_and_Evaluation} under the assumption that disagreement implies that both interpretations cannot hold. Cases where none or only one interpretation is plausible were classified as mistakes.} The analysis was performed on the original data (timeline-based annotations) rather than the data converted to \tlinks, since this allowed us to compare the actual annotations as performed by the annotators without losing any information due to the conversion. We identified 5 main types of variation between the annotators, listed in \autoref{tab:label-variation-table}.

\begin{table}
\setlength{\extrarowheight}{5pt}
    \small	
    \centering
\adjustbox{max width=\textwidth}{    
    \begin{tabular}{cp{1.7cm}p{6.4cm}p{5.2cm}c} 
    \hline
        & \textsc{Category} & \textsc{Description} & \textsc{Example} & \textsc{\%}  \\ \hline
        1. & consecutive vs roughly-simultaneous & While one annotator groups the events together as roughly simultaneous, the other annotates the order explicitly.

        & No one was hurt, but firefighters {\color{purple}[\textit{\textbf{ordered}}]$e_1$} the {\color{purple}[\textit{\textbf{evacuation}}]$e_2$} of nearby homes and said they'll monitor the shifting ground. & 30\% \\ 
        &  &  \textsc{Timeline:} \textbf{{\color{purple}[B]} vs {\color{purple}[B][B]}} on the same temporal position & \textcolor{purple}{[\textit{\textbf{ordered}}]e\textsubscript{1}}  and \textcolor{purple}{[\textit{\textbf{evacuation}}]e\textsubscript{2}} -- consecutive \textit{or} roughly simultaneous 
        &  \\ \hline
        
        2. & different 
        \newline
        positions on 
        \newline
        the timeline & The annotators differ in the way they interpret the event and its temporal position, but both interpretations are plausible. Note that this may also lead to different, yet equally acceptable, annotations of factuality (e.g., interpreting an event as one that did not happen in the past or as one that may happen in the future).

        & 
        (a) Now the ninth US circuit court of appeals has ruled that the original appeal was flawed since it brought up {\color{purple}[\textit{\textbf{issues}}]$e_1$} that had not been raised before. 
        
        (b) The police and prosecutors said they had identified different suspects in six of the cases and had yet to {\color{purple}[\textit{\textbf{find}}]$e_1$} any pattern linking the killings or the victims, several of whom were believed to be prostitutes.
        & 22\% \\ 
        &  &  \textsc{Timeline:} \textbf{{\color{purple}[B]} vs {\color{purple}[B]}} on different temporal positions & \textcolor{purple}{[\textit{\textbf{issues}}]$e_1$} -- when crime was committed \textit{or} when they were brought up

        \textcolor{purple}{[\textit{\textbf{find}}]$e_1$} -- past negated event or future possible event
        &  \\ \hline
        3. & state vs 
        \newline
        action & While one annotator interprets the event as a state that begins at a certain point and lasts through a portion of the story (partially bounded), the other annotates it as a bounded event with the focus being on the action rather than the resulting state. 
     
        & Kidnappers kept their promise to kill a store owner they took hostage and police found the man's {\color{purple}[\textit{\textbf{dismembered}}]$e_1$} and {\color{purple}[\textit{\textbf{decapitated}}]$e_2$} body Friday {\color{purple}[\textit{\textbf{wrapped}}]$e_3$} in plastic garbage bags. & 20\% \\ 
         &  & \textsc{Timeline:} \textbf{{\color{purple}[B]} vs {\color{purple}[U\}/\{U]}} anchored on the same temporal position & \textcolor{purple}{[\textit{\textbf{dismembered}}]$e_1$},  \textcolor{purple}{[\textit{\textbf{decapitated}}]$e_2$}, \textcolor{purple}{[\textit{\textbf{wrapped}}]$e_3$} -- from certain point in the past (state) \textit{or} at certain point in the past (action) &  \\ \hline
        4. & bounded vs centered unbounded  & While one annotator marks an event as bounded, the other treats it as an unbounded event ``centered'' at the same point as the bounded event in the other annotation.%
        This difference in interpretation is common for attitude verbs such as ``think,'' ``hope.'' ``believe.'' 
        
        & And I {\color{purple}[\textit{\textbf{hope}}]$e_1$} that, whatever happens today, that our relationships with Russia will continue to be productive and constructive and strong, because that's very important to the future of our peoples. & 12\%  \\ 
        &  & \textsc{Timeline:} \textbf{{\color{purple}[B]} vs {\color{purple}\{U\}}} on the same temporal position & \textcolor{purple}{[\textit{\textbf{hope}}]$e_1$} -- at the given moment (bounded) \textit{or} overlapping with neighboring events (centered unbounded) &  \\ \hline
        5. & granularity & Annotators' interpretations differ in their level of granularity. These are usually cases where one annotator annotates an unbounded event as permanent/generic, and another annotator adds a ``center'' to that event.
        
        & %
        There have been no {\color{purple}[\textit{\textbf{arrests}}]$e_1$} in any of the slayings.
        & 8\% \\
        &  & \textsc{Timeline:} \textbf{{\color{purple}\{:\}} vs {\color{purple}\{U\}/\{U1:U2\}}} or {\color{purple}\{U1:U2\}} vs {\color{purple}\{U1:U2\}} with a wider interval
        
        & \textcolor{purple}{[\textit{\textbf{arrests}}]$e_1$} -- generally, in the whole story (unbounded) \textit{or} up to the moment of the utterance (centered unbounded)  &  \\ \hline
        6. & mistakes & Any mistake due to honest lapses of judgment. Most mistakes can be attributed to accidentally marking two events or an event or a timex under the same span when that interpretation is impossible or results in other inconsistencies (e.g., marking another event that also relates to the same timex as \textsc{before} or \textsc{after} the event which is already annotated as simultaneous to the times). & ``I haven't seen a pattern yet,'' {\color{purple}[\textit{\textbf{said}}]$e_1$} Patricia Hurt, the Essex County prosecutor, who {\color{purple}[\textit{\textbf{created}}]$e_2$} the task force on Tuesday.  

\vspace{18pt}
One annotator accidentally groups {\color{purple}[\textit{\textbf{said}}]$e_1$} and {\color{purple}[\textit{\textbf{created}}]$e_2$} under one span.       
        & 8\% \\ 
    \end{tabular}
}   
    \caption{Reasons for label variation between the annotators.}
     \label{tab:label-variation-table}
\end{table}

We observe that the biggest single source of label variation stems from the decision to cluster several events together as roughly simultaneous, or  explicitly mark their order (see \textsc{category} 1 in \autoref{tab:label-variation-table}). The resulting timelines are overall comparable, except for the segment in question. Our version of \name\ guidelines embrace this source of variation, allowing the annotators to choose either strategy.  %
It may be possible to train the annotators to be more consistent in this regard.%

We further notice that for some events there may be more than one plausible temporal interpretation: this source of variation corresponds to \textsc{Category} 2 in \autoref{tab:label-variation-table}. Consider the ``issues'' in the example (a). Since they concern a crime, one interpretation is that the issues existed since the crime was committed. Another interpretation is that the issues concern the court case, since that set is not exactly the same as all issues concerning the crime, and in  that case they only exist since the court case.

Note that this kind of difference in temporal perception may also result in varying, yet equally acceptable, annotations of the event factuality. For instance, ``find'' in example (b) can be interpreted as negated event in the past (i.e., ``didn't happen''), or a potential event in the future (i.e., ``maybe will happen'').  %
All examples in this category rely heavily on the annotator interpretation, which can differ due to individual differences as well as cultural background.

An almost equally common reason for label variation is ``state vs action'' (\textsc{Category} 3 in the table): one annotator puts more focus on the underlying action, while the other focuses on the resulting state. This results in annotating the same event as either a bounded event positioned in the past or a partially unbounded event (state) continuing into the future. For instance,  ``decapitated'' ($e_2$) from the example in \autoref{tab:label-variation-table} can be interpreted as a bounded event [B] in the past when the action of decapitation took place, or as the state [U\} resulting from that action, which started at the same moment as the action, but then continued indefinitely into the future. %

The differences in the perceived scope of the event (\textsc{Category} 4) are usually related to attitude verbs, such as ``think'' or ``believe'', which in news texts usually comes in official statements. One possible interpretation is that the attitude is held at the moment of speech, %
in which case they would be annotated as  bounded events ([B]). But it is also plausible that the attitude is held for some time before/after expressing that attitude; in that case they would be annotated as  unbounded  \{U\} events ``centered'' at the moment of speech. 

Finally, we observe some differences due to different granularity of annotation of unbounded events (\textsc{Category} 5). One annotator could interpret an event as a generic/permanent state (unbounded event without a temporal position, encoded as \{:\}), while another could attribute it to a specific period in time + underspecified periods before/after (encoded as \{x\} or \{x:y\}).

Unavoidably, we also find some mistakes, mostly (but not only) due to annotating an event and a timex under the same span. While this annotation is not necessarily problematic, it may lead to errors when there is another event placed before or after the given event that also shares the same time (e.g., both events happen on the same day of the week). Overall, we notice that about 8\% of the difference in annotations can be attributed to mistakes of one or both annotators. This compares to 13\% errors in \textit{all} \tlinks\ in TimeBank 1.2 (an improved version of TimeBank 1.1), reported by %
\newcite{ocal-etal-2022-comprehensive}.\footnote{%
Note that these values (8\% and 13\%) are not directly comparable. In case of \name, the 8\% refers to the 8\% of ``disagreement'' found in the 6 analyzed texts, while in the case of the TimeBank 1.2 the 13\% refers to the 13\% of \textit{all} \tlinks\ (not only disagreement) in the texts  analyzed in \newcite{ocal-etal-2022-comprehensive}.}

\section{Baseline results}

We develop a simple Transformer-based neural relation classification model for \name.
It consists of a LongT5 \cite{Guo2021LongT5ET} encoder and a relation classification head.
We feed a whole TimeBank document into the encoder and then extract contextualized representations of each event and timex into a tensor $H \in \mathbb{R}^{[e \times h]}$. Then, we add a trainable bilinear form to predict relations between every pair of events as $H \cdot W \cdot H^T$, where $W \in \mathbb{R}^{[h \times r \times h]}$, $r$ is the number of relation types and $h$ is the hidden size of LongT5. 

Our choice of LongT5 is motivated by its ability to process long documents, and its availability in different sizes, which allows us to investigate the effect of encoder size on the performance of the full system. Long document support is very important, since most of the annotated documents are more than 400 tokens long (median 442), and some are as long as 2000 tokens. %
To train the network, we used a single A100 40Gb GPU, bf16 precision, and the longest training run took about one hour.

We split the full TimeBankNT corpus into the training set (30 documents) and test set (6 documents), and fine-tune our system on the former. For the test set we select the same six documents that were used in qualitative analysis (\autoref{sec:qualitative}). These documents vary in length, the number of events, and IAA (from $\alpha$=0.47 to $\alpha$=0.85). Since our qualitative analysis (\autoref{sec:qualitative}) indicates that the majority of cases of ``disagreement'' is in fact better described as human label variation. %

The results of our modeling efforts are presented in \autoref{tab:modeling_results}. We performed manual hyperparameter tuning of learning rate and weight decay for each encoder. After initial hyperparameter tuning the variation of accuracy (within a single model) was at most 0.03. Final hyperparameters were: batch size \texttt{32}, learning rate \texttt{1e-4}, weight decay \texttt{0}, dropout \texttt{0.1}. As basic baselines, we used both the most frequent class and a simple rule that assigns events as \textsc{After} if they occur later in the text in relation to other events and \textsc{Before} otherwise. Human results are for one annotator vs. the other.

\begin{figure}
    \centering
    \begin{minipage}[b]{0.4 \linewidth}
        \centering
        \includegraphics[width=\linewidth]{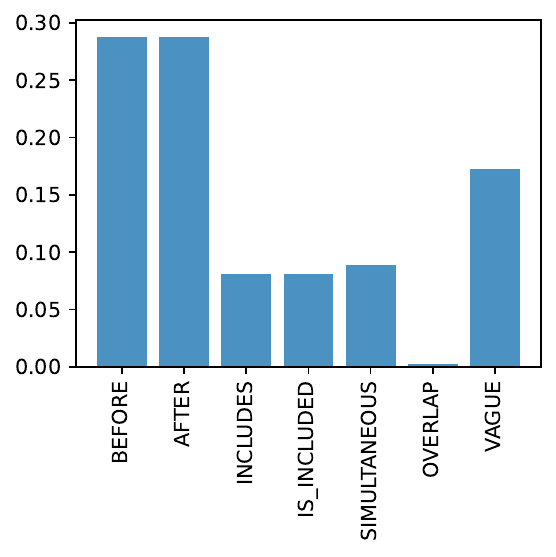}
        \caption{Relation type distribution}
        \label{fig:relation_type_distribution}
    \end{minipage}%
    \begin{minipage}[b]{0.5\linewidth}
        \centering
        \includegraphics[width=\linewidth]{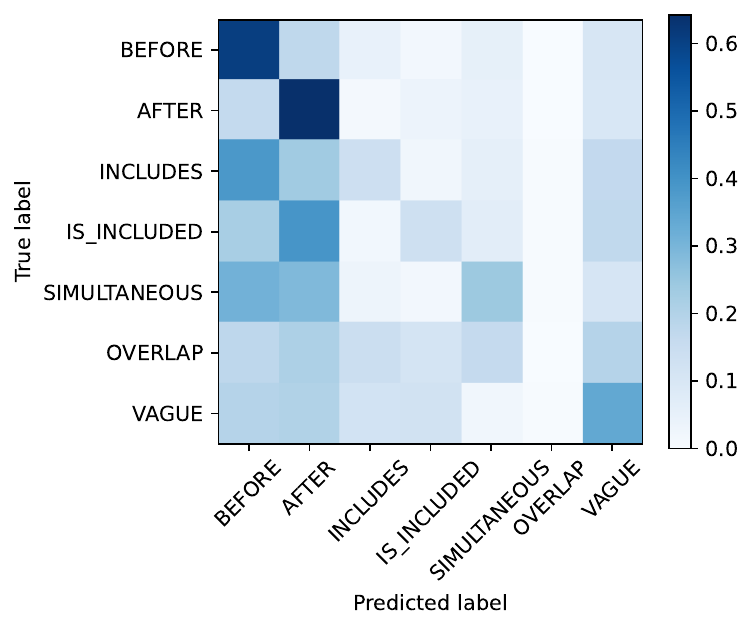}
        \caption{Relation prediction confusion matrix}
        \label{fig:confusion_matrix}
    \end{minipage}
\end{figure}

Even the best model only reaches F1 of 0.31, which shows that the task is quite challenging. One possible reason for that is the inclusion of \textit{all} relations between all events and all timexes, which includes both short- and long-distance relations. Additionally, \autoref{fig:relation_type_distribution} shows significant imbalance in the distribution of relation data, with the \textsc{Overlap} relation being the least frequent at only 0.2\%. At the same time, our simple ``later is after'' heuristic baseline achieves only 30\% accuracy (and much lower F1), which shows that the temporal structure of these news texts is indeed much more complex.

\begin{table}[t]
    \centering
    \begin{tabular}{l|c|c|c|c}
        \toprule
         & \textsc{Accuracy} & \textsc{Precision} & \textsc{Recall} & \textsc{F1} \\
        \midrule
        Most frequent class & 0.30 & 0.04 & 0.15 & 0.07 \\
        ``Later is after'' heuristic & 0.30 & 0.09 & 0.14 & 0.11 \\
        \midrule
        LongT5 Base (114M) & 0.44 & 0.32 & 0.29 & 0.29 \\
        LongT5 Large (349M) & 0.45 & 0.35 & 0.28 & 0.29 \\
        LongT5 XL (1253M) & 0.47 & 0.34 & 0.31 & 0.31 \\
        \midrule
        Human performance & 0.73 & 0.58 & 0.59 & 0.57 \\
        \bottomrule
    \end{tabular}
    \caption{Modeling results. Precision, recall, and F1 are macro-averaged over relation types.}
    \label{tab:modeling_results}
\end{table}

The confusion matrix for our best model configuration (\autoref{fig:confusion_matrix}) shows that the model overpredicts frequent \textsc{Before} and \textsc{After} relations (especially at the expense of \textsc{Simultaneous}), and almost never predicts the rare \textsc{Overlap} relations. Interestingly, the asymmetrical relations \textsc{Before} and \textsc{After} seem to be confused with another asymmetrical relations pair \textsc{Includes} and \textsc{Is\_included}.

\begin{figure}
    \centering
    \includegraphics[width=\textwidth]{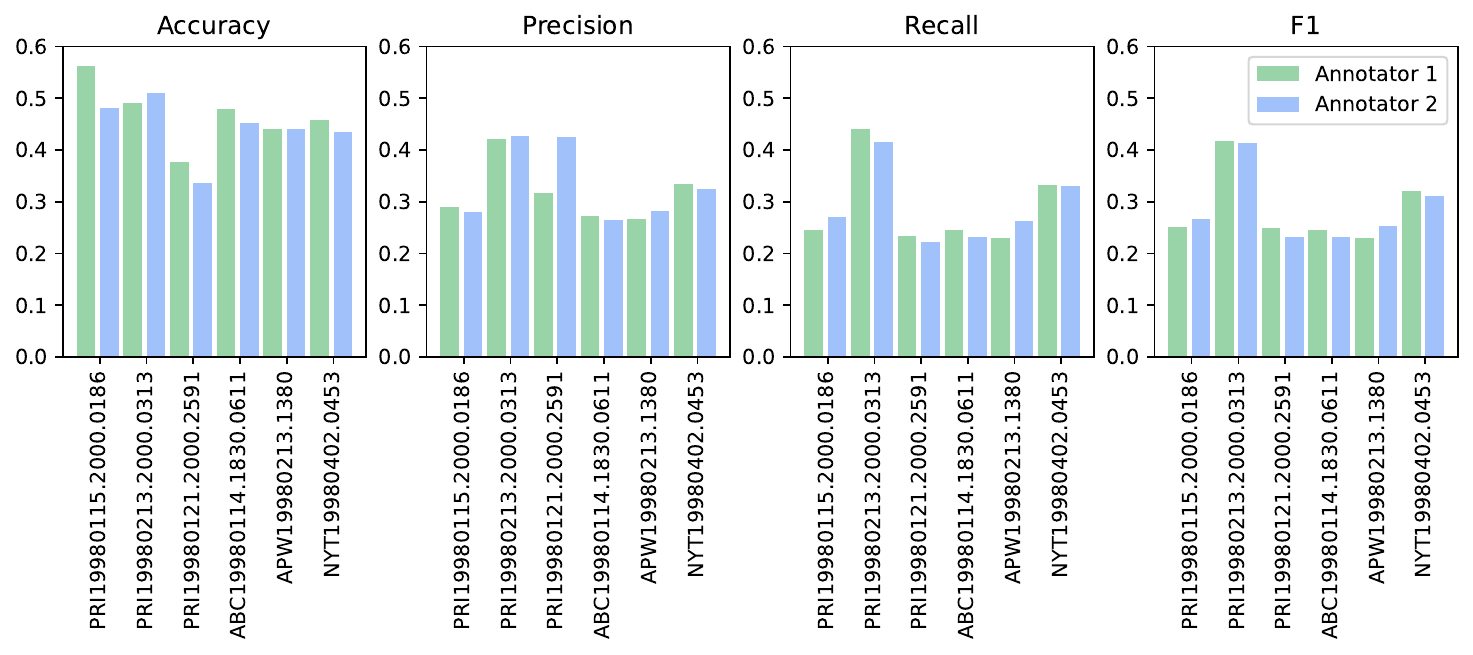}
    \caption{Per-document metrics}
    \label{fig:per_doc_metrics}
\end{figure}

Looking at the per-document metrics (\autoref{fig:per_doc_metrics}) we observe that the system does not rely excessively on either of the annotators. In the case of \texttt{PRI19980115.2000.0186}, it could be related to the ``consecutive vs roughly simultaneous'' human label variation case (row 1 in \autoref{tab:label-variation-table}). %
The \texttt{NYT19980402.0453} document is interesting because the IAA for it is low ($\alpha$=0.47), but the model's accuracy remains similar for both of the annotators.

\paragraph{Long-distance relations vs. short-distance relations}
Since temporal relations between long-distance events are a distinctive feature of \name, we perform an additional evaluation on events that are closer than ten words (roughly adjacent sentences) vs. further than 100 words. \autoref{tab:short_rels_vs_long_rels} shows that both of the relations are hard to model, as the model makes more mistakes in these two classes than in general (\autoref{tab:modeling_results}). This suggests that medium-distance (10-100 words) relations are the simplest to predict. Low numbers on close-by event relations can also be explained by the confusion between \textsc{Simultaneous} and \textsc{Before}/\textsc{After} (\autoref{fig:confusion_matrix}), which in turn could be partly due not to errors, but to the label variation in consecutive vs roughly-simultaneous case (row 1 in \autoref{tab:label-variation-table}). %

\begin{table}[ht]
    \centering
    \begin{tabular}{l|c|c|c|c}
        \toprule
         & \textsc{Accuracy} & \textsc{Precision} & \textsc{Recall} & \textsc{F1} \\
         \midrule
         All events & 0.47 & 0.34 & 0.31 & 0.31 \\
         \midrule
         Closeby events & 0.27 & 0.17 & 0.23 & 0.19 \\
         Far events & 0.41 & 0.32 & 0.28 & 0.29 \\
         \bottomrule
    \end{tabular}
    \caption{Short-distance and long-distance relations metrics.}
    \label{tab:short_rels_vs_long_rels}
\end{table}

\paragraph{A note on large language models.} 

In-context learning with large language models (LLMs) is an increasingly popular approach, but it is challenging to test on temporal relation classification with \name\ data for two reasons. First, the high density of \tlinks\ means that 
a single generation cannot produce all thousands or tens of thousands of relations that are typically present in a \name\ document, since the input length is usually limited to 2048 or 8192 tokens. Generating relations one by one is prohibitively expensive, especially via a paid API. Even in the case of the publicly available models \cite{Scao2022BLOOMA1,Zhang2022OPTOP} a lot of time on expensive hardware is required in order to evaluate a single document.

The second problem is specific to our choice of TimeBank data. Since it has been a very popular dataset, it is highly likely that large 
language models have seen this data coupled with temporal annotations during their pre-training, especially if the models were trained on unfiltered GitHub repositories. This would constitute a test set leak. %

\section{Future work}

The \name\  improvements in temporal annotation density and handling of underspecification open up several exciting prospects for future work.

\paragraph{More data with dense temporal annotation.} By enabling dense temporal annotation at a fraction of the cost of full manual annotation with traditional event pairs, \name\ provides means to create new resources for training ML models and more challenging benchmarks, including in particular long-distance temporal relations \cite{NaikBreitfellerEtAl_2019_TDDiscourse_Dataset_for_DiscourseLevel_Temporal_Ordering_of_Events}.

\paragraph{Fine-grained vagueness.} A general problem with prior sparse approaches is telling \textit{why} no temporal relation exists between a given pair of events: did the annotator just not consider it, or considered it and decided that no relation exists, or that multiple relations are all possible \cite{chambers2014dense}? \name\ solves this problem by (a) ensuring that annotator does explicitly consider every possible relation by putting everything on a timeline, (b) providing three mechanisms for handling different cases of underspecification: different timeline branches, unbounded events, factuality values.

Since \name\ explicitly distinguishes between temporal order underspecification due to unbounded events vs different timeline branches or factuality, these cases can now be targeted for additional commonsense reasoning annotation and inference \cite{ZhouKhashabiEtAl_2019_Going_on_vacation_takes_longer_than_Going_for_walk_Study_of_Temporal_Commonsense_Understanding}. For example, in a sentence \textit{John woke up, went to work, got off the bus, came to the office, stopped his podcast.} we don't know exactly when he started listening to the podcast, but we know it probably did include the bus time because people often listen to podcasts when they commute. Given \name\ annotation, we would be able to tell when the model should try to reason about likely event duration.

\paragraph{The death of the ``gold standard''?} This work showed a significant amount of genuine variation in temporal annotation \autoref{sec:qualitative}, which reinforces the need to move away from the traditional ``gold standard" approach to temporal annotation \cite{Plank_2022_Problem_of_Human_Label_Variation_On_Ground_Truth_in_Data_Modeling_and_Evaluation}. Rather than trying to adjudicate such cases, we need to start modeling the possible interpretations by different people. We release our own TimeBankDense corpus, fully double-annotated, and we hope that \name\ framework would enable more such resources in the future.

\paragraph{More qualitative analysis.} By re-annotating the same TimeBankDense corpus that was already used in a number of temporal annotation projects \cite[inter alia]{PustejovskyHanksEtAl_2003_The_TIMEBANK_Corpus,CassidyMcDowellEtAl_2014_An_Annotation_Framework_for_Dense_Event_Ordering,NingWuEtAl_2018_A_Multi-Axis_Annotation_Scheme_for_Event_Temporal_Relations,NaikBreitfellerEtAl_2019_TDDiscourse_Dataset_for_DiscourseLevel_Temporal_Ordering_of_Events}, we enable a comparative study on inter-annotator agreement between these projects, which all used different guidelines, different sets of TimeML relations, different annotation setups. Let us stress that \name\ is an annotation \textit{framework} primarily innovating in the mechanisms to improve annotation density and handling of underspecification. In principle, it could be coupled with different temporal annotation \textit{schemes} with respect to definitions of events, temporal expressions, factuality etc., and all of these could impact IAA.

\paragraph{Human-friendly annotation.} Another key innovation of \name\ is the span-based approach to annotation that allows placing whole clusters of events on a single timeline point. Since this approach is arguably more psychologically realistic (see \autoref{sec:cog}), it also brings up the question to what degree the natural reading chunking strategies map onto \name\ annotation behavior, and whether it works differently for different annotators. It also brings up the question of how to compute IAA in frameworks that allow annotations that are different on the surface, but semantically equivalent. %

\section{Conclusion}

We present \name, a new framework for temporal annotation that is based on interactive timeline representation of the whole text rather than ordering individual event pairs.  \name\ achieves IAA comparable or superior to the prior art, but offers significantly denser annotation, three mechanisms for handling underspecification, and support for a more natural reading process. We contribute \name\ guidelines, open source tools for annotation and conversion to standard TimeML format, as well as TimeBankNT corpus: the densest TimeBank, with 36 documents each annotated by two expert annotators. 

We also conduct a detailed qualitative analysis of 6 complete TimeBank texts, which suggests that most disagreements are in fact best described as legitimate human label variation, and the field needs to embrace learning from such data rather than aim for a single ``gold standard''.

\bibliography{bibtex}
\bibliographystyle{coling}

\clearpage

\appendix

\end{document}